\documentclass[a4paper,11pt]{article}

\usepackage{PRIMEarxiv}
\usepackage[utf8]{inputenc}
\usepackage[T1]{fontenc}
\usepackage{hyperref}
\usepackage{url}
\usepackage{booktabs}
\usepackage{amsfonts}
\usepackage{nicefrac}
\usepackage[disable]{microtype}
\usepackage{lipsum}
\usepackage{fancyhdr}
\usepackage{graphicx}
\graphicspath{{image/}}
\usepackage{float}
\usepackage{xcolor}
\usepackage{amsmath,amsthm}
\usepackage{subcaption}
\usepackage{array}
\usepackage{longtable}
\usepackage{ragged2e}
\usepackage{geometry}
\usepackage{enumitem}
\usepackage{setspace}

\usepackage{wrapfig}
\definecolor{cite_color}{rgb}{0.0, 0.58, 0.71}
\definecolor{headerBlue}{RGB}{51,122,183}
\definecolor{lightGray}{RGB}{245,245,245}
\definecolor{borderGray}{RGB}{220,220,220}
\definecolor{headerColor}{RGB}{60,60,60}
\definecolor{lineColor}{RGB}{200,200,200}

\usepackage[square,numbers,sort&compress]{natbib}
\setcitestyle{numbers} 

\hypersetup{
    colorlinks  = true,
    citecolor   = cite_color,
    linkcolor   = cite_color
}

\renewcommand{\figurename}{Fig.}
\makeatletter
\renewcommand*{\fnum@figure}[1]{\figurename~\thefigure.}
\makeatother

\renewcommand{\arraystretch}{1.3}
\def\tsc#1{\csdef{#1}{\textsc{\lowercase{#1}}\xspace}}
\tsc{WGM}
\tsc{QE}
\tsc{EP}
\tsc{PMS}
\tsc{BEC}
\tsc{DE}

\usepackage{amsmath} 
\usepackage{amssymb} 
\usepackage{mathtools} 

\title{OpenLKA: an open dataset of lane keeping assist from market autonomous vehicles
}
\vspace{-30pt} 
\author{
    Yuhang Wang$^{1}$, 
    Abdulaziz Alhuraish$^{1}$, Shengming Yuan$^{1}$, Shuyi Wang$^{2}$, Hao Zhou$^{1}$\thanks{Corresponding author: \texttt{haozhou1@usf.edu}}  \\[6pt]
    \small
    $^{1}$Civil \& Environmental Engineering, University of South Florida, Tampa, FL, USA \\
    $^{2}$Department of Transportation Engineering, Fuzhou University, Fuzhou, China
}

\begin{document}
\maketitle

\vspace{-40pt} 
\begin{figure}[htbp]
  \centering
  \includegraphics[width=0.8\textwidth]{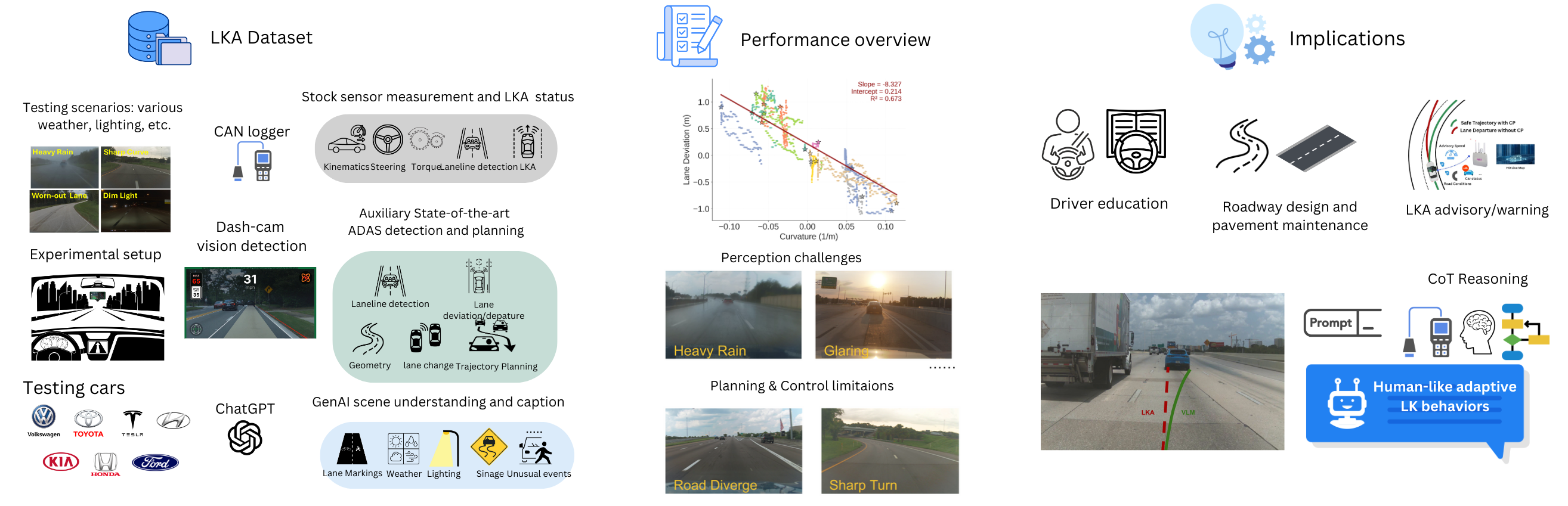} 
  \caption{OpenLKA: A comprehensive framework for LKA analysis and enhancement.}
  \label{fig:overview}
\end{figure}

\begin{abstract}
The Lane Keeping Assist (LKA) system has become a standard feature in recent car models on the road. While marketed as providing auto-steering capabilities, the system's operational characteristics and safety performance remain under explored, primarily due to a lack of real-world testing and comprehensive data. 
To address the gap, we conducted extensive testing of mainstream LKA systems from leading automakers in the U.S. market during the summer of 2024 in Tampa, Florida. Employing an innovative experimental method, we collected a comprehensive dataset that includes full Controller Area Network (CAN) messages with LKA attributes, as well as video, perception, and lateral trajectory planning data captured using a high-quality front-facing camera equipped with advanced vision detection and trajectory planning algorithms. Testing scenarios encompassed diverse and challenging conditions such as complex road geometry, adverse weather, varying lighting, degraded lane markings, and combinations of these factors. A vision language model (VLM) is utilized to further annotate the videos including features of weather, lighting, and surrounding traffic conditions etc. Based on the dataset gathered, this paper provides an empirical overview of the system's operational features and safety performance. Key findings indicate that: (i) LKA's perception is susceptible to faint lane markings and low contrast against pavement, reducing its effectiveness; (ii) LKA struggles in lane transition areas, such as merges, diverges, and intersections, often resulting in unintended lane departures or disengagements; (iii) LKA's steering torque limitations contribute to frequent lane deviations or departures on sharp turns, posing significant safety risks; and (iv) LKA systems consistently exhibit a rigid lane-centering tendency, lacking the adaptability to adjust lane position on sharp curves or near large vehicles such as trucks.  The paper concludes by highlighting how the dataset can serve both the infrastructure and self-driving technology sectors. In light of LKA limitations, we provide recommendations for improving road geometry design and pavement maintenance practices. Additionally, we demonstrate how the dataset can support the development of human-like LKA systems through the application of VLM with fine-tuning and Chain of Thought reasoning. The code and data are shared at: \url{https://github.com/OpenLKA/OpenLKA}.
\end{abstract}


\section{Introduction}

The proliferation of Advanced Driver Assistance Systems (ADAS) in modern vehicles marks a pivotal shift in automotive technology \cite{nidamanuri2021progressive, yurtsever2020survey}. Among these, Lane Keeping Assist (LKA) \cite{lazcano2021mpc} has become a standard feature in recent car models, autonomously modulating steering to maintain lane centering \cite{monfort2022speeding, dean2022estimating}. Despite its widespread adoption, LKA systems' real-world performance limitations \cite{waghchoure2022comprehensive} create critical challenges across multiple stakeholders. For drivers, the uncertainty about LKA's capabilities \cite{rizgary2024driver} leads to confusion about when and where to engage the system, resulting in both steep learning curves and potential safety risks. Infrastructure operators and transportation authorities lack clear guidelines for adapting road design elements \cite{gopalakrishna2021impacts}, such as lane marking specifications and departure warning systems, to better support LKA functionality. Meanwhile, the autonomous driving technology sector faces obstacles in improving these systems due to limited sharing of performance data and failure cases within the research community, hindering collaborative efforts to enhance LKA reliability and safety \cite{orieno2024future}.

While manufacturers rigorously test their systems during development in controlled environments, these evaluations fail to capture the complexity of real-world driving conditions. Some transportation departments have conducted preliminary assessments of road infrastructure impact on LKA performance \cite{pike2024assessing}, their studies remain limited in scope, focusing on specific scenarios with a small number of test vehicles. The lack of comprehensive performance data across various vehicle models and manufacturers has left critical questions unanswered about LKA's real-world capabilities, particularly under challenging conditions such as complex road geometries, adverse weather, and deteriorated lane markings \cite{becker2020safety}. This scarcity of extensive, multi-modal datasets has impeded both systematic assessment of existing systems and development of more robust algorithms \cite{ye2020automated, duan2020hierarchical, makridis2021openacc}.

To address these challenges, we introduce OpenLKA, a large-scale open dataset of LKA systems on market car models.  While preliminary efforts have been made to assess LKA performance in limited scenarios \cite{sultana2021lane, nguyen2021testing}, our study represents a significant advancement in both scope and methodology. Since its launch in the summer of 2024, this systematic data collection initiative combines high-fidelity data from vehicle CAN networks, advanced vision systems, and machine learning techniques within a unified framework. Through extensive tests conducted in Tampa, Florida, we have leveraged the region's subtropical climate to evaluate LKA performance under diverse weather patterns and road configurations. Additionally, we have incorporated driving data from the open-source community, significantly enhancing our dataset's diversity and providing valuable insights into how different driving styles interact with LKA systems. This multi-faceted approach not only captures basic vehicle control data but also rich contextual information about the driving environment, enabling a thorough evaluation of LKA performance under real-world conditions.

Our comprehensive analysis reveals significant limitations in current LKA systems across perception, planning, and control modules. While these systems perform adequately under normal conditions, they exhibit considerable deviations from lane center or even disengage completely when encountering scenarios such as sharp curves \cite{chang2024evaluation} or deteriorated lane markings. More importantly, compared to human drivers' adaptability and safety awareness, LKA systems demonstrate rigid lane-centering behavior \cite{mueller2024habits}, potentially leading to hazardous situations in edge cases. This inflexibility stands in stark contrast to human drivers' ability to adjust their lateral position based on road conditions, surrounding vehicles, and other contextual factors.

The key contributions of our paper are:

\begin{enumerate}[label=\Roman*.]
\item We present OpenLKA, an extensive multi-modal driving dataset collected from a diverse fleet of LKA-equipped vehicles. Our data collection framework combines reverse-engineered OBD-II interfaces for accessing LKA system states, high-resolution dash-cam recordings, and state-of-the-art lane detection algorithms. The dataset encompasses over 130 hours of driving data across 15 different vehicle models, supplemented by comprehensive scene annotations generated through vision-language models, providing detailed interpretations of traffic scenarios, environmental conditions, and driving contexts.

\item Through systematic overview of OpenLKA, we categorize and quantify LKA limitations across perception, planning, and control modules. Our empirical investigation reveals significant performance degradation in challenging scenarios, particularly on curved road segments. We establish quantitative relationships between lane-keeping deviations and road geometry parameters such as curvature radius and rate of curvature change, supported by extensive real-world measurements and statistical evidence.

\item Based on our empirical findings, we develop concrete recommendations for improving LKA technology and infrastructure readiness. For LKA developers, we demonstrate how vision-language models can enhance system adaptability by incorporating human-like reasoning capabilities in lane-keeping decisions. For transportation authorities, we provide specific guidelines for lane marking designs and curve advisories that better support LKA operations. Additionally, we propose a standardized framework for evaluating and benchmarking LKA performance, facilitating more transparent development and validation processes within the autonomous driving community.
\end{enumerate}

We present an overview of our complete pipeline in Fig.~\ref{fig:overview}. The insights and methodologies developed through this research not only advance our understanding of current LKA systems but also establish a foundation for developing more adaptive and robust autonomous driving technologies that better align with human driving behavior.

\section{The dataset overview}
\label{sec:dataset}

\begin{figure}[htbp]
  \centering
  \includegraphics[width=0.8\textwidth]{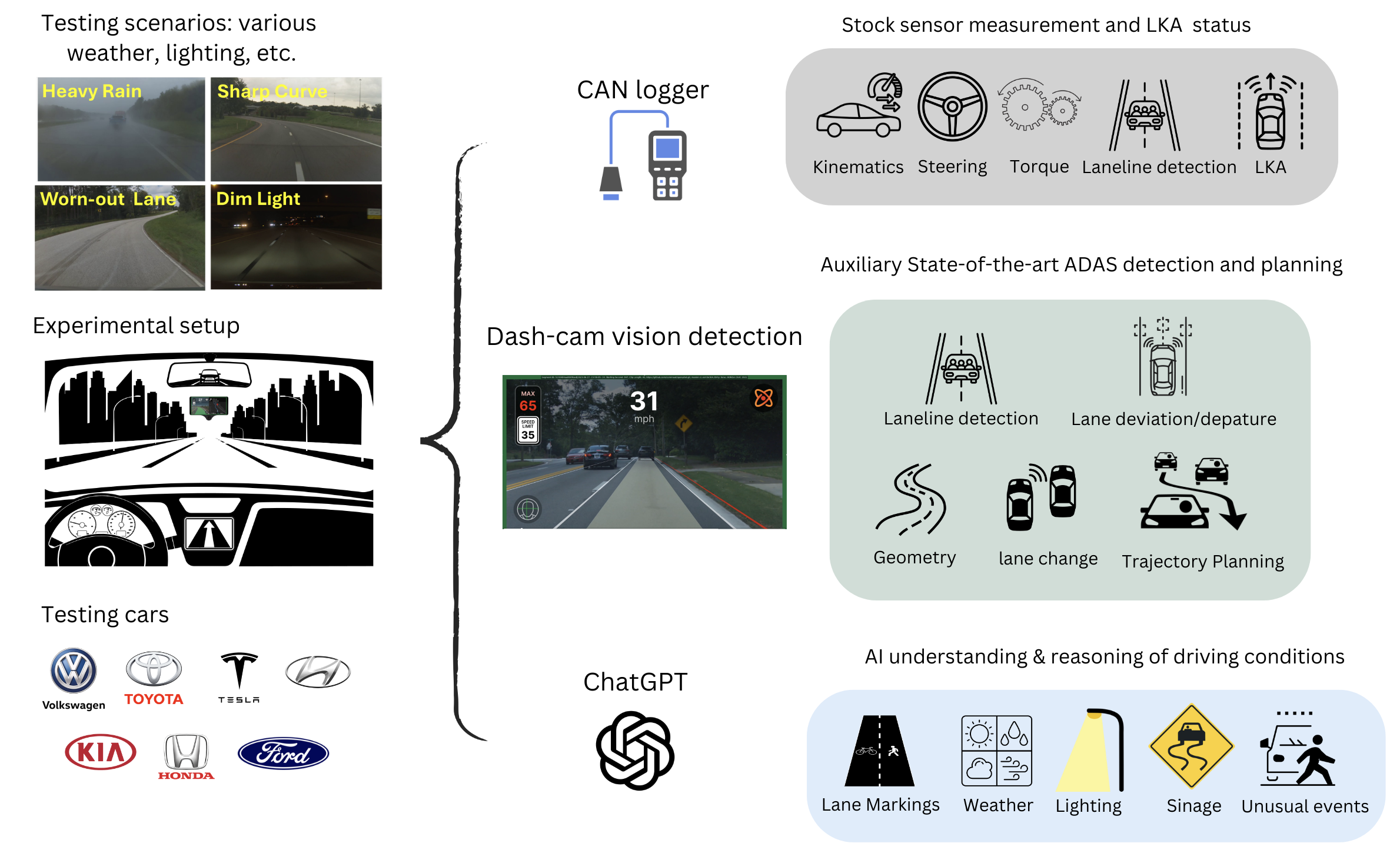} 
  \caption{LKA Dataset Overview}
  \label{fig:lka_overview}
    \vspace{-1em}
\end{figure}

The OpenLKA dataset is composed of two main components. The first and primary component is the LKA dataset, which comprises vehicle data collected from more than a dozen different models operating under LKA conditions. The brands of these vehicles represent over 80\% of the current market share, providing a broad overview of LKA performance in commercially available vehicles. This dataset contains over 100 hours of driving data under LKA conditions, offering a comprehensive resource for analysis. Figure \ref{fig:lka_overview} provides an overview of the LKA dataset. A portion of the data comes from Openpilot by Comma.ai, where data related to vehicle control was captured via the Harness connection to the vehicle's CAN bus while the device was in dash camera mode. Openpilot provides the vehicle's current trajectory, the trajectory predicted by an end-to-end model for the subsequent 10 seconds, and the distance to the edges of the roadway along this predicted path. Since Openpilot is vision-based, the dataset also includes information about the speed of one or two preceding vehicles and their distance from the subject vehicle. In addition, another part of the LKA dataset originates from CAN messages captured from different vehicles. Due to the diversity in CAN formats across vehicle models, reverse engineering was conducted on each of them, leading to variations in the variables included for different vehicles. The bottom half of Figure~\ref{fig:lka_overview} illustrates data collected from different vehicle models in this manner.

\begin{figure}[htbp]
  \centering
  \includegraphics[width=0.6\textwidth]{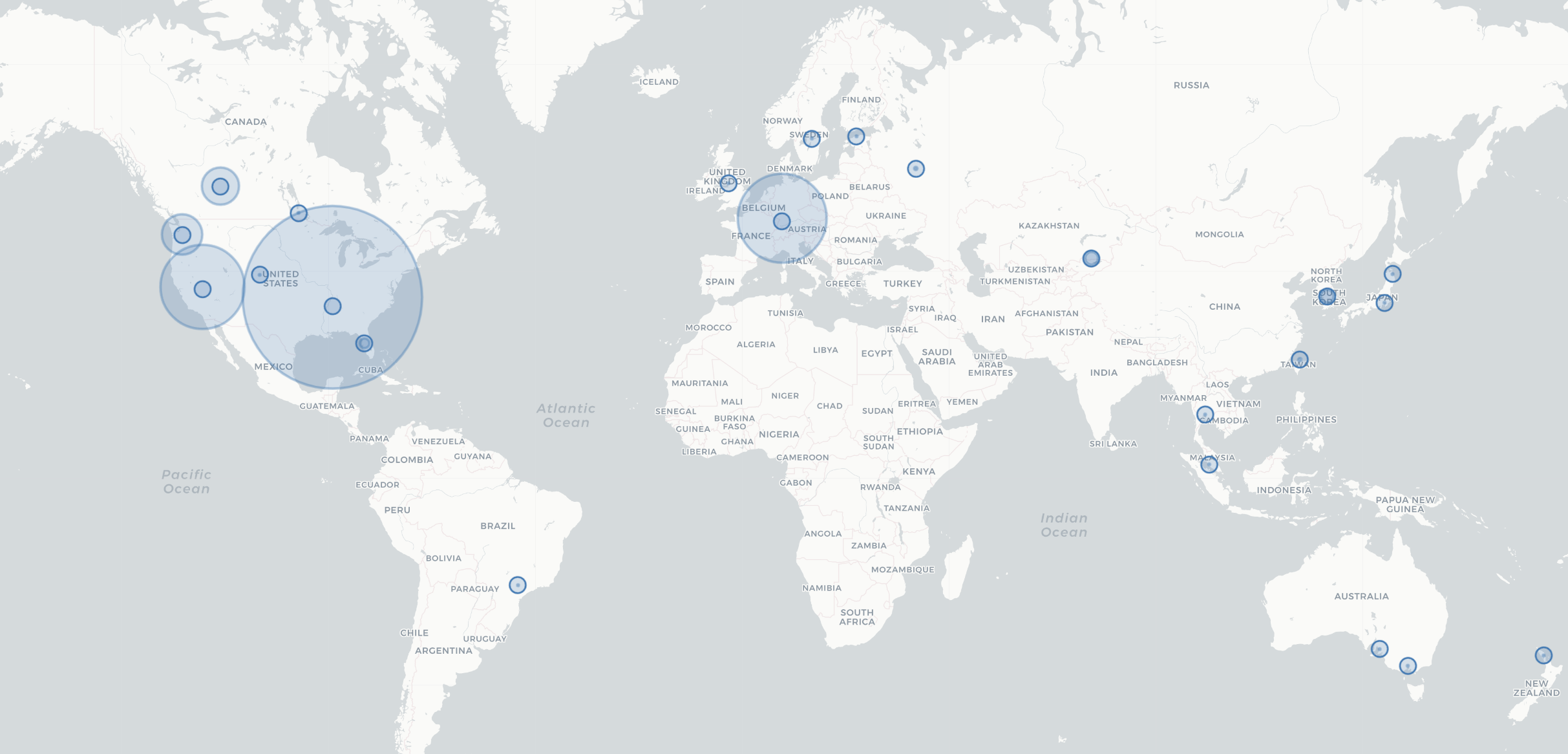}
  \caption{Human Dataset from Drivers Across the World}
  \label{fig:hv_overview}
    \vspace{-1em}
\end{figure}

The second part of the OpenLKA dataset features human driving data collected as both a comparison and supplement to the LKA dataset. Through the Open Comma platform, we gathered over 1,300 driving segments from more than 280 drivers worldwide. These human driving logs are compiled using Openpilot’s qlog, consistent with the Openpilot data part structure illustrated in Figure~\ref{fig:hv_overview}. Because human drivers demonstrate more flexible decision-making and control over the vehicle, modeling their behavior can help the LKA system make more intuitive and safer decisions. Moreover, drawing on human driving styles allows us to customize the LKA’s driving characteristics, thereby enhancing both safety and enjoyment in autonomous driving.

In addition, this dataset includes driving videos synchronized with the vehicle control data, available in .ts and .hevc formats, capturing both the wide-angle front view and the windshield view. Compared to many existing autonomous driving datasets, OpenLKA offers comprehensive multimodal information. Leveraging these videos, we enhance the dataset using a large language model to extract additional parameters that are not directly obtainable from CAN messages or Openpilot data, such as weather conditions, traffic congestion levels, lighting conditions, and road markings. This makes OpenLKA one of the most diverse and parameter-rich LKA datasets available. The table below presents a comparison between OpenLKA and other self-driving datasets, highlighting the significant advantages of our dataset and showcasing its potential for fostering new avenues of dataset-driven research.

\begin{table}[htbp]
\caption{Comparison of OpenLKA Dataset with Existing Datasets}
\label{tab:comparison}
\centering
\footnotesize
\setlength{\tabcolsep}{5pt}  
\begin{tabular}{@{}p{2.2cm}|p{0.7cm}p{0.7cm}p{0.7cm}|p{0.7cm}p{0.7cm}p{0.7cm}p{0.7cm}p{0.7cm}p{0.7cm}|p{0.7cm}p{0.9cm}@{}}
\toprule
\textbf{Dataset} & \multicolumn{3}{c|}{\textbf{Data}} & \multicolumn{6}{c|}{\textbf{Env Factors}} & \textbf{LLM} & \textbf{Time} \\
\cmidrule(lr){2-4} \cmidrule(lr){5-10} \cmidrule(lr){11-12}
 & Vis & CAN & LiD & Wthr & Vis & Traf & Infra & Evnt & Lght & Ann & (hrs) \\
\midrule
\textbf{OpenLKA} & \textbf{1} & \textbf{1} & \textbf{0} & \textbf{1} & \textbf{1} & \textbf{1} & \textbf{1} & \textbf{1} & \textbf{1} & \textbf{1} & \textbf{150+} \\
HDD & 1 & 0 & 0 & 1 & 1 & 1 & 1 & 1 & 1 & 0 & 104 \\
BDD-X & 1 & 0 & 0 & 1 & 1 & 1 & 1 & 1 & 1 & 1 & 77 \\
WoodScape & 1 & 0 & 0 & 1 & 1 & 1 & 1 & 1 & 1 & 0 & N/A \\
DAWN & 1 & 0 & 0 & 1 & 1 & 1 & 1 & 1 & 1 & 0 & N/A \\
KITTI & 1 & 0 & 1 & 0* & 0 & 0 & 0 & 0 & 0 & 0 & 1.5 \\
BDD100K & 1 & 0 & 0 & 1 & 1 & 1 & 1 & 1 & 1 & 0 & 1.2K \\
Comma2k19 & 1 & 1 & 0 & 0 & 0 & 0 & 0 & 0 & 0 & 0 & 33 \\
Cityscapes & 1 & 0 & 0 & 1 & 1 & 1 & 1 & 1 & 1 & 0 & N/A \\
ApolloScape & 1 & 0 & 1 & 1 & 1 & 1 & 1 & 1 & 1 & 0 & N/A \\
nuScenes & 1 & 0 & 1 & 1 & 1 & 1 & 1 & 1 & 1 & 0 & 1.2K+ \\
Waymo & 1 & 0 & 1 & 1 & 1 & 1 & 1 & 1 & 1 & 0 & 570+ \\
\bottomrule
\end{tabular}
\caption*{\small *Limited weather conditions in KITTI dataset.}
\end{table}

\section{Experiments and Data Processing}
\label{sec:experiments}

\subsection{Experiment Methods}
\label{subsec:methods}

\begin{wrapfigure}{r}{0.45\textwidth}
  \vspace{-1em}
  \centering
  \includegraphics[width=\linewidth]{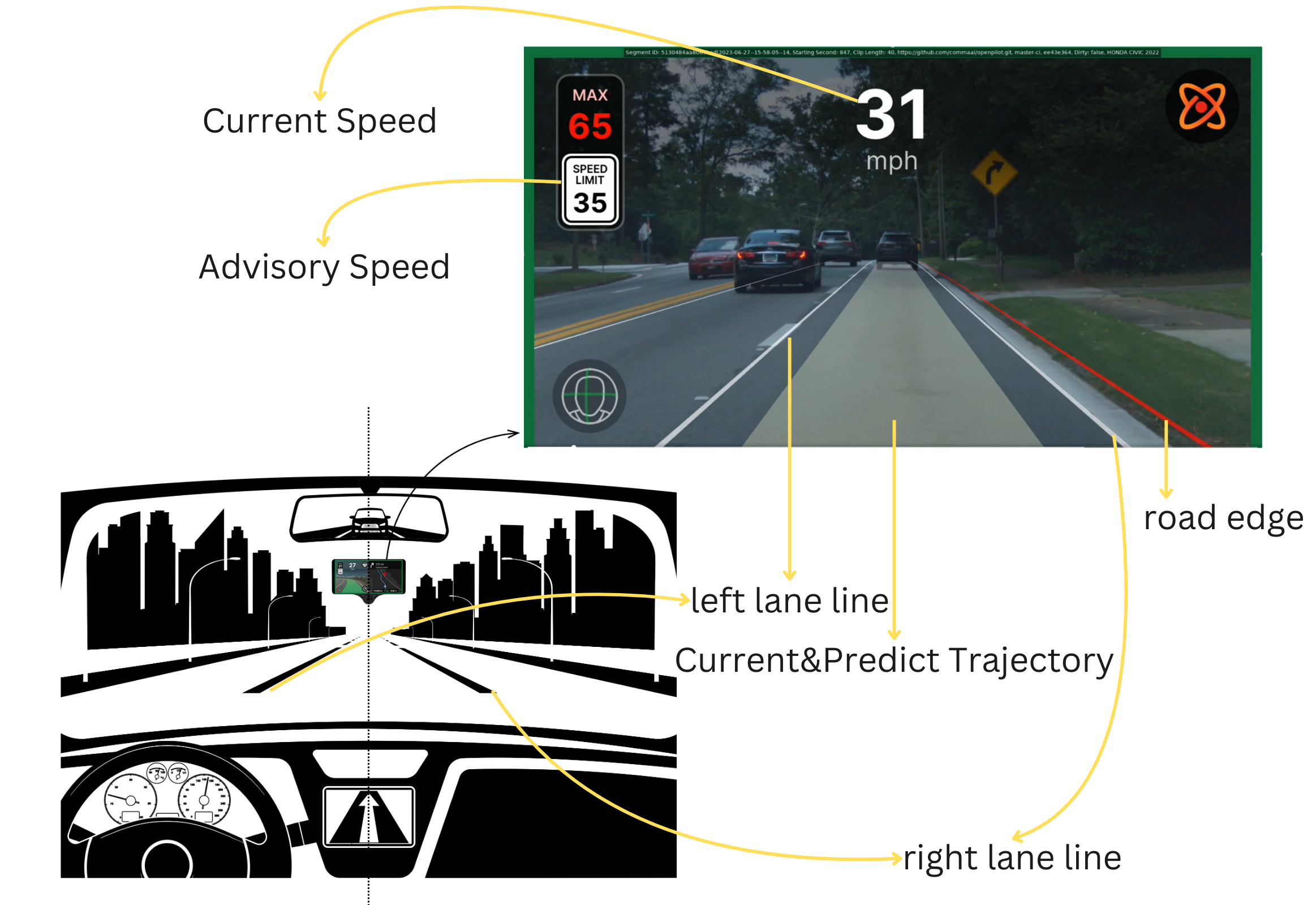}
  \caption{Comma 3x, dash cam for LKA data collection, placed at the front windshield of the vehicle.}
  \label{fig:comma_setup}
  \vspace{-0.3em}
\end{wrapfigure}

To empirically evaluate the real-world performance \cite{re2021testing} of LKA Systems in commercially available vehicles \cite{song2023identifying}, we employed advanced data collection equipment designed to capture a comprehensive set of vehicle dynamics and environmental parameters. The primary tool used was the Comma Three-X \cite{comma_three_x}, an advanced aftermarket self-driving development kit tailored for research purposes. This device seamlessly integrates with a vehicle's systems, enabling the recording of Controller Area Network (CAN) messages from the vehicle’s On-Board Diagnostics (OBD-II) interface, as well as capturing high-definition front-view video through an integrated dash camera.

We mounted the Comma Three-X device at the central position of the vehicle's interior windshield, shown in Fig. ~\ref{fig:comma_setup}, to simplify the calculation of lane deviation. This central placement ensured an optimal field of view for the device's cameras and sensors, providing accurate and consistent data collection across all test vehicles. The Comma Three-X is equipped with a sophisticated lane detection algorithm that leverages computer vision and machine learning techniques to accurately identify lane lines in real time. This capability allowed us to calculate the vehicle’s lateral position within the lane, estimate lane-centering errors, and detect instances of lane departure. By synchronizing CAN data with video footage, we achieved a holistic view of the vehicle's mechanical responses in relation to external driving environments.

To ensure that the vehicle's native control systems remained unaffected during data collection, we utilized the Comma Openpilot in \emph{dash-cam mode}. In this mode, the device does not exert any control over the vehicle but passively reads data from the vehicle's CAN bus. This approach enabled us to gather detailed insights into the driving environment and the vehicle's control signals without interfering with the stock LKA system. By monitoring the vehicle's sensor readings and control commands, we could assess the performance of the LKA system under various real-world conditions while maintaining the integrity of the vehicle's original functionalities.

\begin{figure}[htbp]
  \centering
  \includegraphics[width=0.8\textwidth]{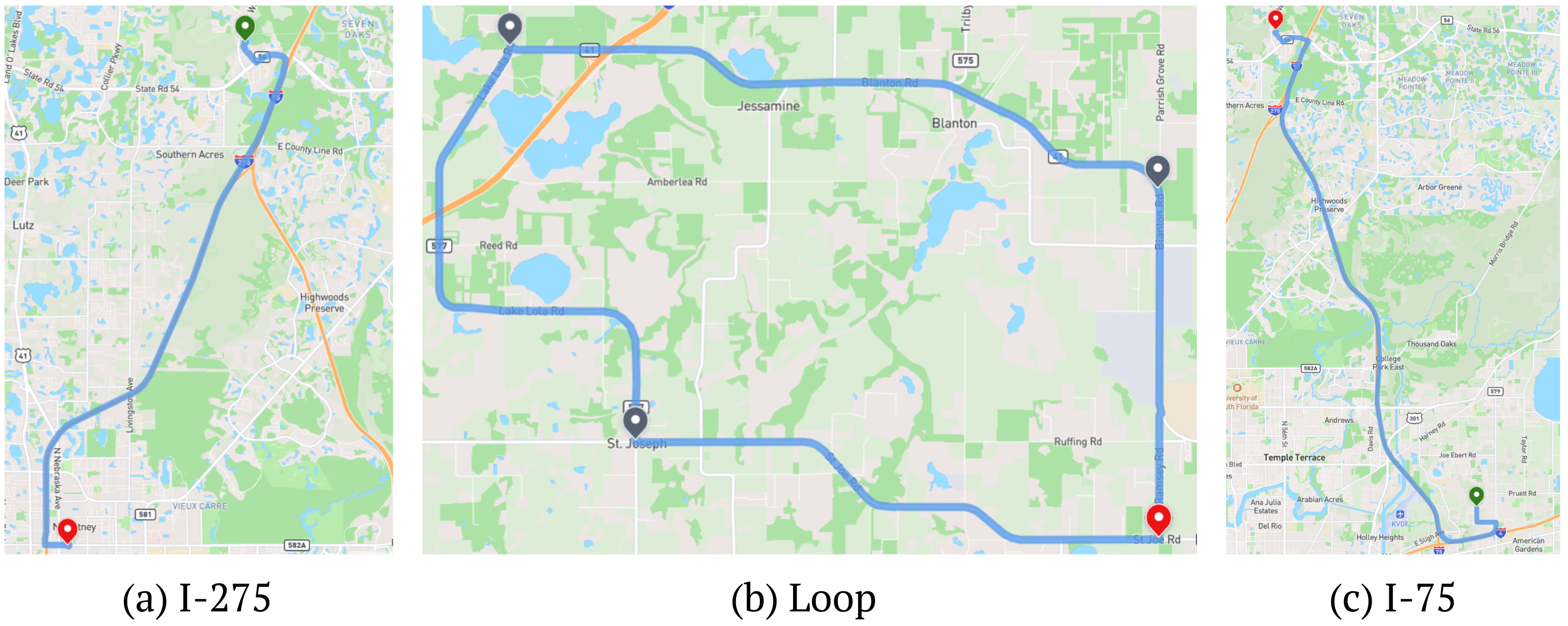}
  \caption{LKA data collection sites}
  \label{fig:layouts}
\end{figure}

\subsection{Testing scenarios}
\label{subsec:scenarios}

\begin{wrapfigure}{r}{0.36\textwidth}
  \centering
  \vspace{-2em} 
  \includegraphics[width=0.35\textwidth]{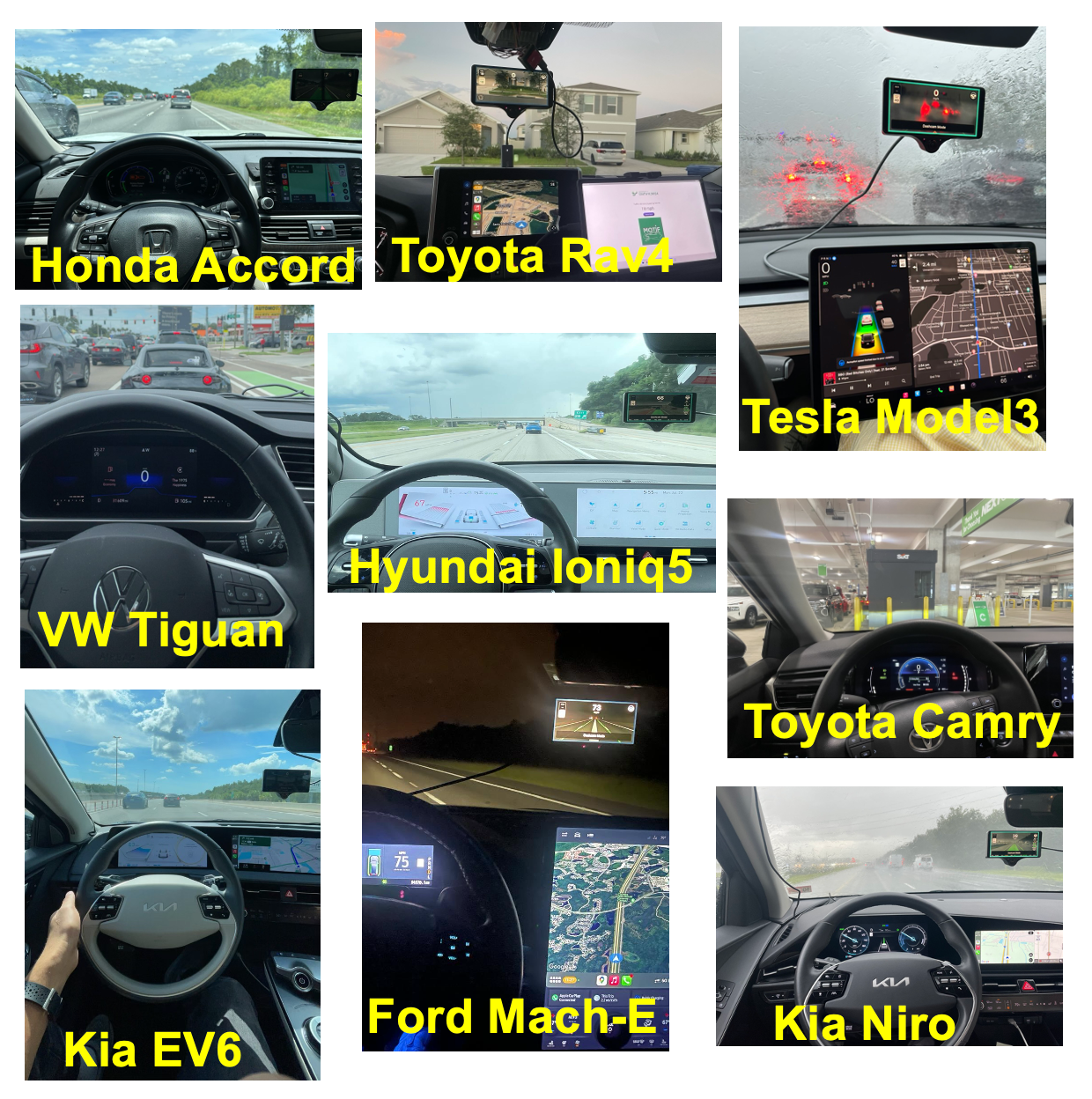}
  \caption{Test Vehicle Fleet Distribution and Market Share Analysis}
  \vspace{-2em}
  \label{fig:Exp_cars}
\end{wrapfigure}

Our experimental design incorporated an extensive array of driving scenarios to rigorously assess LKA performance under real-world conditions. We conducted tests using a diverse fleet of rental cars sourced from Tampa International Airport, including models from major automakers such as Honda, Toyota, Tesla, Ford, Kia, and Hyundai. This diversity ensured a broad representation of LKA technologies currently accessible to consumers, capturing a wide spectrum of implementations and performance characteristics crucial for a generalizable assessment across different manufacturers and models.

We meticulously considered a wide range of variables to capture the behaviors of LKA in different environments, as shown in Table~\ref{tab:testing_conditions}, the subtropical climate of Tampa provided an ideal setting to test under adverse conditions, such as sharp curve, heavy rain and high humidity—common challenges for LKA performance. By intentionally selecting complex and demanding road segments, we aimed to expose potential limitations and failure modes inherent in the systems.

The test drives were conducted by experienced drivers thoroughly trained in the operation of the LKA systems and the data collection equipment. Safety precautions were rigorously observed; drivers remained attentive and prepared to assume control of the vehicle whenever necessary. This approach ensured that we gathered valuable data on LKA performance while maintaining the highest safety standards throughout the study.

\begin{table}[h]
\caption{Summary of Testing Variables and Conditions}
\label{tab:testing_conditions}
\begin{tabular}{@{}p{4cm}p{11cm}@{}}
\toprule
\textbf{Variable} & \textbf{Conditions Tested} \\
\midrule
\textbf{Weather Conditions} & Clear skies; Rain; High humidity; Heavy thunderstorms \\
\addlinespace
\textbf{Lighting Conditions} & Daytime; Nighttime; Dawn; Dusk; Sudden changes due to thunderstorms; Glare from sunlight; Street lighting variations; Tunnel transitions \\
\addlinespace
\textbf{Pavement Quality} & Smooth asphalt; Rough surfaces; Potholes; Uneven lanes; Wet surfaces; Gravel roads; Dirt roads; Recently repaired patches \\
\addlinespace
\textbf{Lane Markings} & Well-defined; Faded; Obscured; Absent markings; Double lines; Dashed lines; Temporary construction markings; Reflective markers \\
\addlinespace
\textbf{Road Geometries} & Straightaways; Sharp curves; Upgrades; Downgrades; Complex intersections; Construction zones; Lane merges/splits; Inconsistent signage; Unconventional lane configurations; Roundabouts; Bridges; Tunnels \\
\addlinespace
\textbf{Traffic Conditions} & Light traffic; Moderate traffic; Heavy congestion; Stop-and-go traffic \\
\addlinespace
\textbf{Vehicle Speeds} & Low speed (<20 mph); Medium speed (20–45 mph); High speed (>45 mph); Variable speeds due to traffic; Sudden acceleration/deceleration \\
\addlinespace
\textbf{Driver Behaviors} & Steady driving; Frequent lane changes; Sudden braking; Rapid acceleration; Distracted driving simulations; Manual override of LKA \\
\bottomrule
\end{tabular}
\end{table}

\subsection{CAN log data acquisition and post-processing}
\label{subsec:can_processing}

Using the Comma Three-X device to access CAN logs, we employed open database container (DBC) files from a variety of market-available vehicles to reverse engineer and extract specific CAN messages related to vehicle lateral control. This reverse engineering process involved collecting open-sourced DBC files, systematically identifying the required variables in OpenDBC, and aligning and validating these variables against the log files in our dataset. Through meticulous validation, we organized all the necessary CAN variables. These variables, detailed in the central section of Figure 1, include timestamps for synchronizing with Openpilot, vehicle lateral and longitudinal control data, and ADAS-related information such as road and surrounding object detection. This rigorous approach ensures that the OpenLKA dataset captures precise and comprehensive vehicle information, setting a high standard for lateral control research.

\begin{figure}[h]
  \centering
  \includegraphics[width=\textwidth]{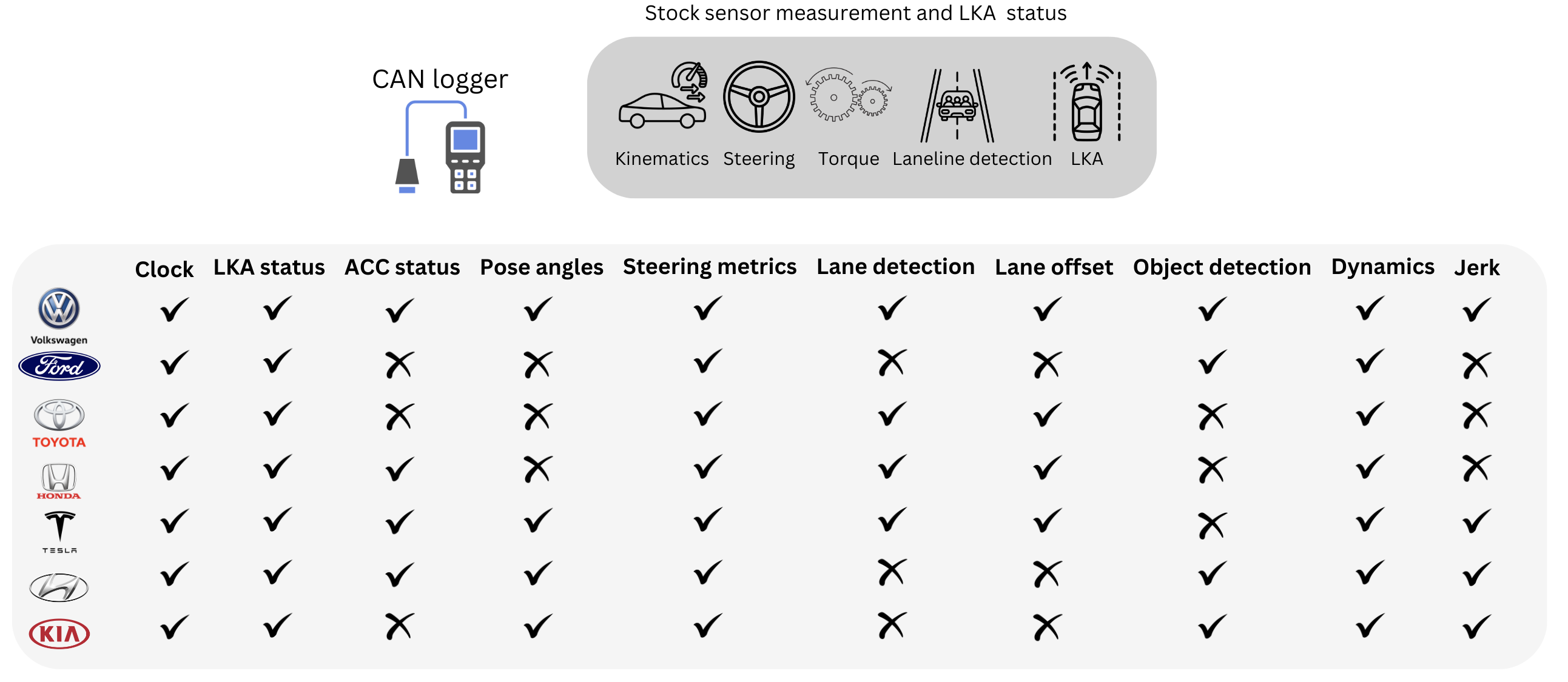}
  \caption{Variables collected through CAN from various car brands}
  \label{fig:CANM}
\end{figure}

The OpenLKA dataset is a multimodal resource that integrates both log data from CAN and visual data recorded by the Comma device. The video data provides valuable contextual insights, enabling qualitative assessments of external factors that influence LKA performance. By combining visual and CAN data, we analyzed key metrics such as lane departures, lane-centering deviations, and the vehicle's lateral positioning within the lane. This holistic approach enhances the understanding of LKA behavior under varying real-world conditions.

For raw data post-processing, we synchronized the CAN data and video footage to accurately correlate the vehicle's mechanical responses with the driving environment. Advanced data analysis techniques—including statistical analyses and machine learning algorithms—were employed to identify patterns, trends, and anomalies in LKA performance across different scenarios.

Over the course of the study, we amassed a dataset exceeding 7,000 kilometers of driving distance and over 130 hours of operational data, encompassing more than 200 individual trips. This extensive dataset not only ensured a robust performance assessment but also enhanced the statistical significance of our findings, allowing for meaningful comparisons and trend analyses across various conditions and vehicle models.

Utilizing a comprehensive experimental setup combined with meticulous data acquisition and post-processing methods, valuable insights are delivered into the real-world performance of LKA in consumer vehicles. The high granularity of our data not only aids in planning processes but also assists in training more accurate feedforward dynamics models. These findings carry significant implications for automakers, policymakers, and researchers in autonomous driving technologies, contributing to the refinement of LKA systems, informing regulatory standards, and ultimately enhancing the safety and efficiency of road transportation.

\subsection{Video-based data augmentation using Large Vision-Language Model}
\label{subsec:video_augmentation}

\begin{figure}[htbp]
  \centering
  \includegraphics[width=0.7\textwidth]{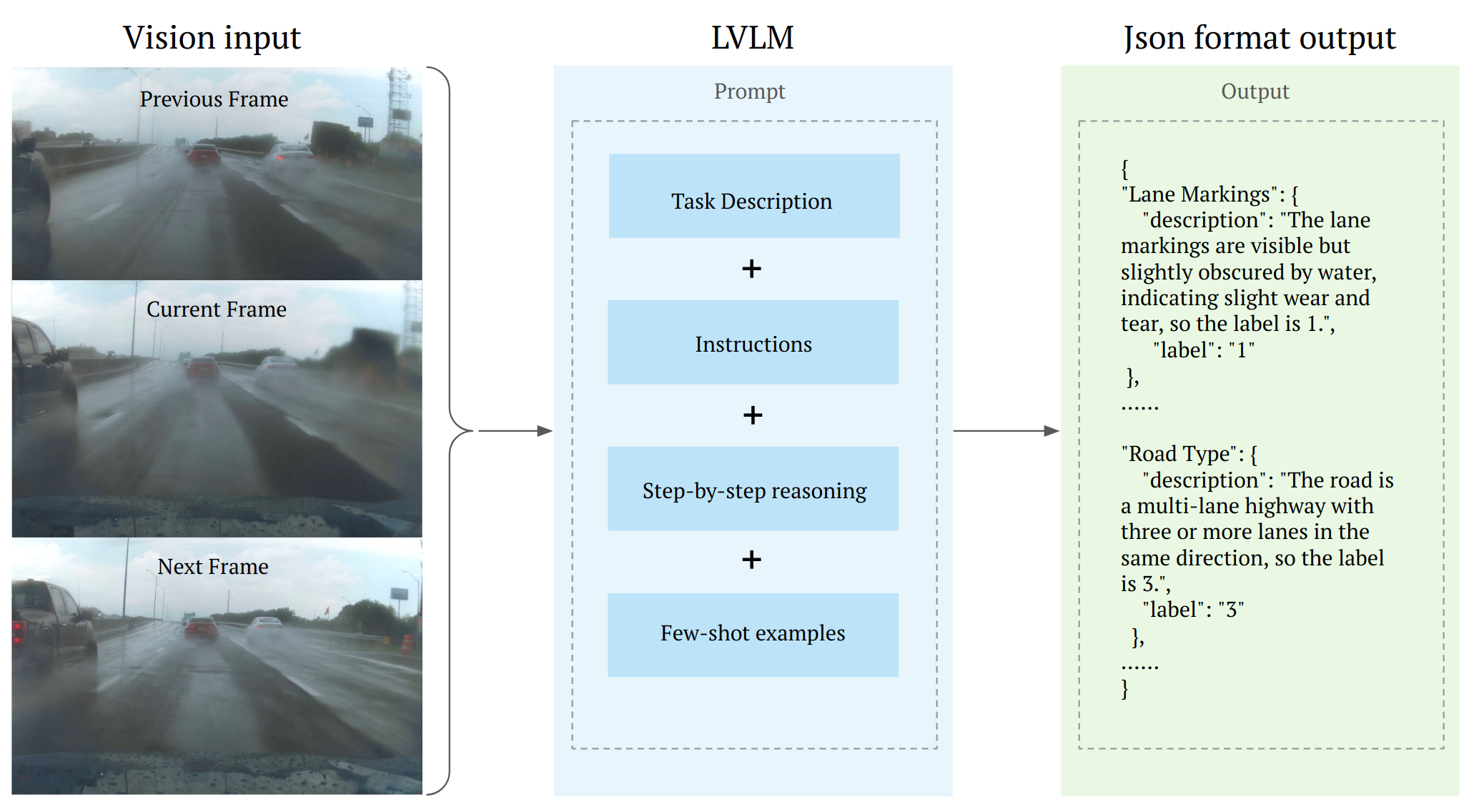}
  \caption{VLM data annotation based on video images}
  \label{fig:llm_annotation}
\end{figure}

After organizing the aligned CAN data and vehicle video footage, we enhanced the OpenLKA dataset to better support the development of various machine learning models by leveraging a large-scale vision-language model (LVLM) to enrich the dataset based on the video data. We draw on mainstream LLMs methods \cite{tan2024large, yang2024data} to assist in manual annotation and find that by using the right Prompts for guidance, the Generalized Knowledge Model GPT-4o visual model already has a good capability in handling the annotation of simple autonomous driving tasks.

Specifically, we first divided the video data into individual video frames. Considering the placement of the Comma device, and to avoid obstructions in certain scenarios caused by windshield wipers or water droplets on the windshield, we included contextual information by pairing each video frame with its preceding and subsequent frames. This approach allowed the LVLM to understand the frames in a clearer and more coherent context. Moreover, we designed a refined prompt tailored for the OpenLKA dataset, incorporating concise and precise language with domain expertise in traffic flow. By integrating the "Chain-of-Thoughts" methodology, this prompt significantly improved the accuracy of LVLM annotations, as illustrated in Appendix~\ref{tab:prompt-structure}.

\begin{table}[htbp]
\centering
\caption{Description of labels}
\label{tab:label_description}
\footnotesize  
\renewcommand{\arraystretch}{1.2}  
\setlength{\tabcolsep}{2pt}  
\begin{tabular}{@{}p{1.7cm}cccccccc@{}}
\toprule
\textbf{Category} & \textbf{0} & \textbf{1} & \textbf{2} & \textbf{3} & \textbf{4} & \textbf{5} & \textbf{6} \\ \midrule
\textbf{Lane Mark.} & Disappear & Unclear marks & Intact marks & - & - & - & - \\ 
\textbf{Weather} & Rain/moisture & Dusty & Clear & Snowy & - & - & - \\ 
\textbf{Lighting} & Dark & Artif. lighting & Normal light & - & - & - & - \\ 
\textbf{Traffic} & Traffic jam & Stop and go & Congested flow & Free flow & - & - & - \\ 
\textbf{Road Cond.} & Large potholes & Small cracks & Smooth road & - & - & - & - \\ 
\textbf{Driving} & Reckless & Signif. anom. & Minor anom. & Normal & - & - & - \\ 
\textbf{Pedestrian} & No pedestrians & Few ped. & Moderate act. & Crowded & - & - & - \\ 
\textbf{Visibility} & Poor vis. & Moderate vis. & Clear vis. & - & - & - & - \\ 
\textbf{Road Type} & Two-lane & Three-lane & Four-lane & Multi-lane & Roundabout & Intersect. & - \\ 
\textbf{Scenarios} & None & Oncoming veh. & Emerg. veh. & Road const. & Obstacle & Animal & Accident \\ 
\textbf{Surr. Veh.} & No vehicles & Low density & Mod. density & High density & - & - & - \\ 
\bottomrule
\end{tabular}
\end{table}

With the assistance of the LVLM, we enriched the OpenLKA dataset in several dimensions (see Table~\ref{tab:prompt-structure} in Appendix). This includes obtaining additional perspectives on the external environment, such as visual perception from the driver’s viewpoint \cite{zhou2022incorporating}, traffic flow details, and road facility information—insights often absent in many existing autonomous driving datasets. These enhancements empower autonomous driving systems to consider a broader array of perspectives and external factors during design and training, thereby improving system robustness.

To validate the reliability of our methods, and considering the inherent similarity of scene types within the OpenLKA dataset, we randomly selected video samples corresponding to different LKA factors. We used our domain knowledge in Transportation to annotate over 2,000+ images from these samples under a unified standard. These annotations were then compared to those generated by GPT-4o (Vision), revealing an accuracy rate of over 95\% for the model. Upon examining the discrepancies in annotations, we found that GPT-4o (Vision) consistently avoided overlooking hazardous situations. On the contrary, its annotations demonstrated a higher level of rigor, often classifying scenarios as more severe. For instance, as shown in Figure X, during a rainy day, water splashes from a vehicle ahead partially obscured the road. While experienced human drivers might consider the lane markings clearly visible, LVLMs conservatively flagged the splashes as a potential interference with lane detection, reflecting this caution in its annotations.

For augmented data aimed at designing safety alert systems in machine learning, this conservative labeling is not only reasonable but also advantageous. By emphasizing potential risks, such annotations contribute to the development of safer, more reliable autonomous driving systems.

\begin{figure}[htbp]
  \centering
  \includegraphics[width=0.8\textwidth]{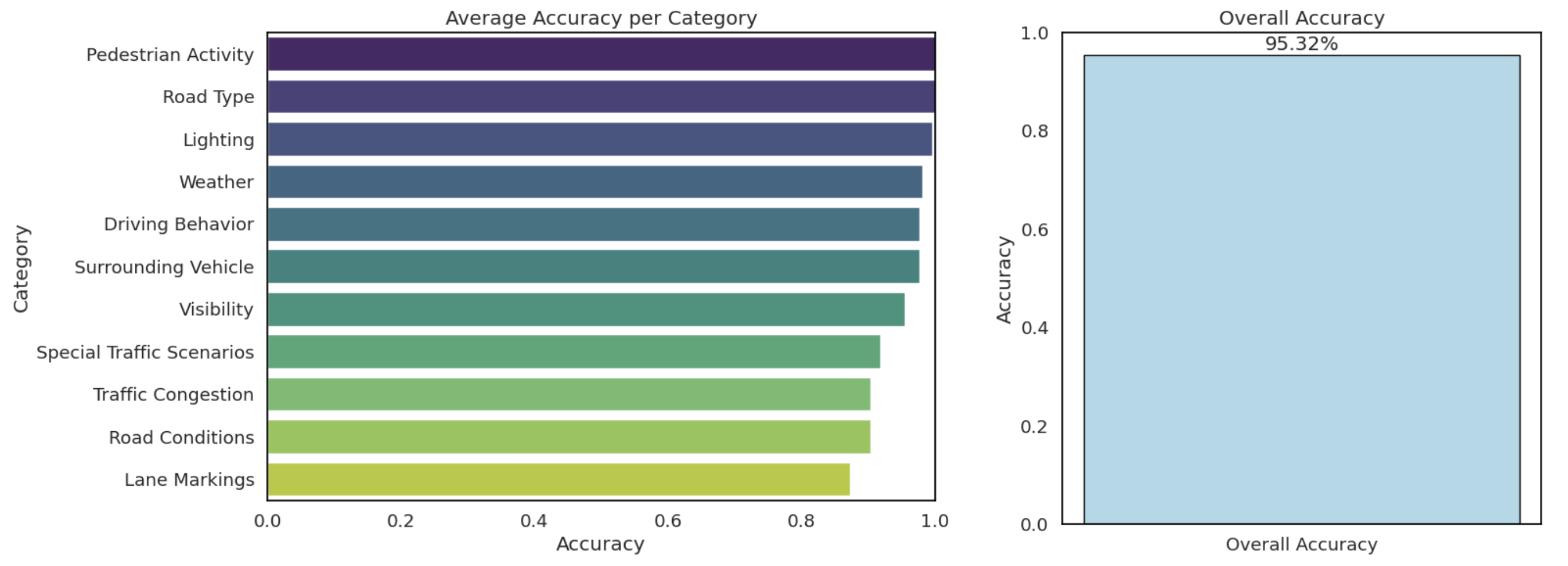}
  \caption{Evaluation result of VLM annotation}
  \label{fig:overall_result}
\end{figure}

\section{Overview of LKA Characteristics and Performance Limitations}
\label{sec:overview}

In this section, we first evaluated the lane deviation and control stability of the LKA system under normal driving scenarios. Subsequently, we analyzed challenging scenarios for LKA, categorizing and examining them based on the root causes across perception, planning, and control domains. Subsequently, we identified scenarios with potentially hazardous implications for LKA systems under specific combined factors and provided a detailed analysis of these risks. Finally, we make a comparision of LKA driving and Human driving and point out the lack of current LKA systems.

\subsection{LKA Characteristics Under Normal Conditions}
\label{subsec:normal_conditions}

We used two metrics to evaluate the lateral stability. First, we show the steering angle profile, which can also be used to calculate the actual curvature of the moving trajectory and the lateral acceleration of the car. Second, we show the LKA error, which is the the distance from the middle of the car to the center of the lane, also called lane deviation. We find that in normal conditions, LKA performance is reliable, which is in line with our expectations.

\subsubsection{Lateral stability in normal conditions}
\label{subsubsec:lateral_stability}

\begin{figure}[htbp]
  \centering
  \includegraphics[width=\textwidth]{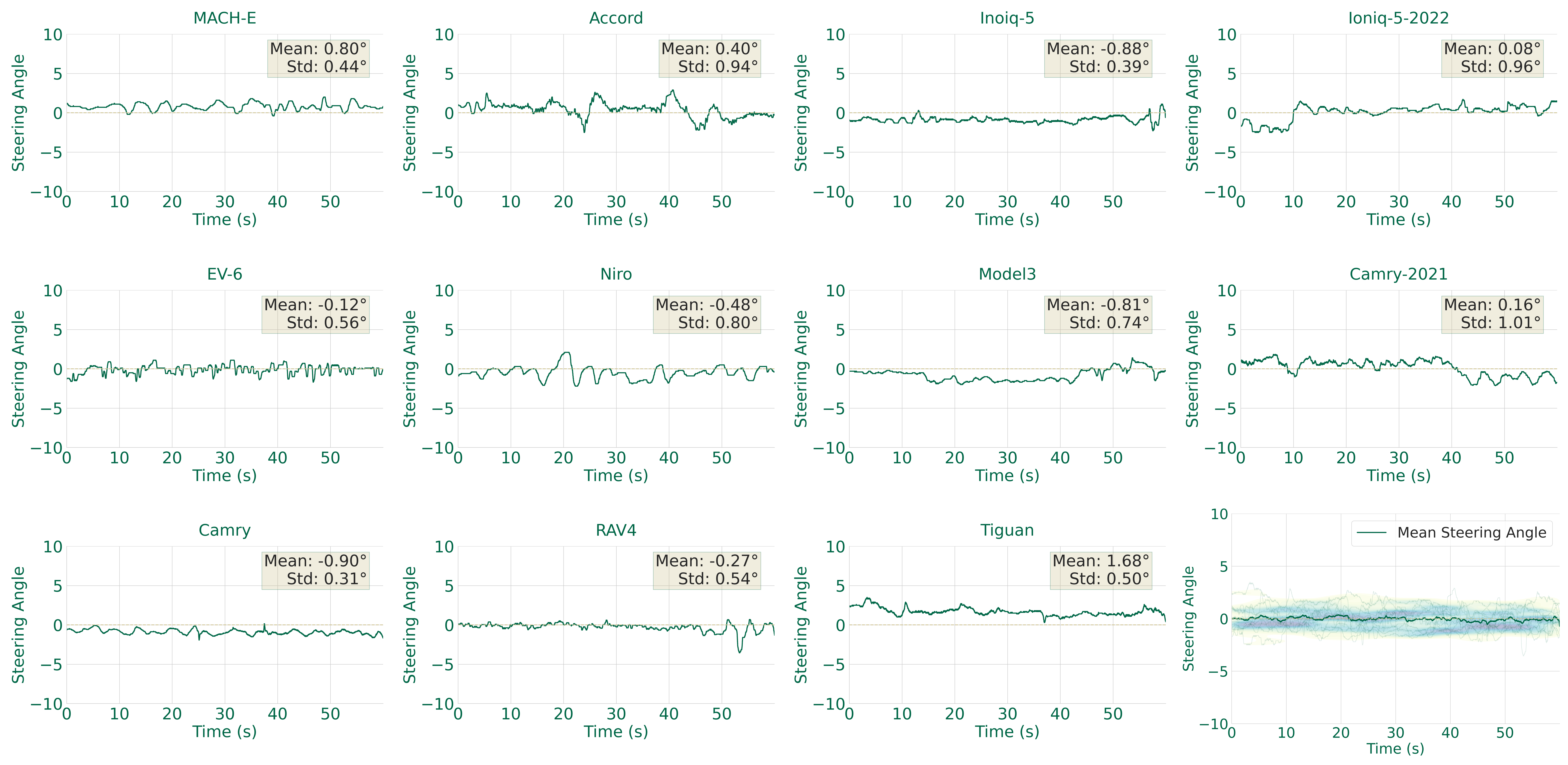}
  \caption{Steering angle of Different Cars in normal conditions}
  \label{fig:steering_angle_overview}
\end{figure}

The steering angle is defined as the angle between the orientation of the front wheels and the vehicle's longitudinal axis. This parameter is fundamental in vehicle dynamics, directly influencing the vehicle's trajectory and handling characteristics \cite{cui2021path}. By adjusting the steering angle, drivers control the direction of the vehicle and maintain a desired path on the road. In LKA systems \cite{Mukherjee2024}, steering angle serves as a valuable metric \cite{stubler2021development, park2021toward} for evaluating system stability and performance. Although LKA systems may employ various methods for lateral control—such as steering angle control, steering torque control \cite{Becker2021}, or a combination \cite{vignati2022cooperative} thereof—the resultant changes in steering angle reflect the system's efforts to maintain lane position. 

A stable LKA system should exhibit smooth and minimal steering angle adjustments \cite{liu2022real} to keep the vehicle centered within the lane without causing discomfort to the driver or passengers. Excessive or erratic steering angle variations may indicate instability or inefficiency in the LKA control algorithms.

Fig.~\ref{fig:steering_angle_overview} illustrates the steering angle behavior over time for various vehicle models operating under normal driving conditions. The observed fluctuations around zero degrees indicate that the LKA systems are making minor adjustments to maintain lane position, thereby demonstrating stable performance. The small magnitude of these changes over time suggests that LKA systems can sustain lane-centering stability over extended periods, enhancing driving comfort by reducing the need for frequent or significant steering interventions. Furthermore, this consistent fluctuation near zero indicates that current LKA systems rely on fixed rules, which are relatively rigid in centering the vehicle, rather than adapting to multiple environmental factors.

\subsubsection{Lane Deviation in normal conditions}
\label{subsubsec:lane_deviation}

\begin{wrapfigure}{r}{0.4\textwidth}
  \centering
  \vspace{-3em} 
  \includegraphics[width=\linewidth]{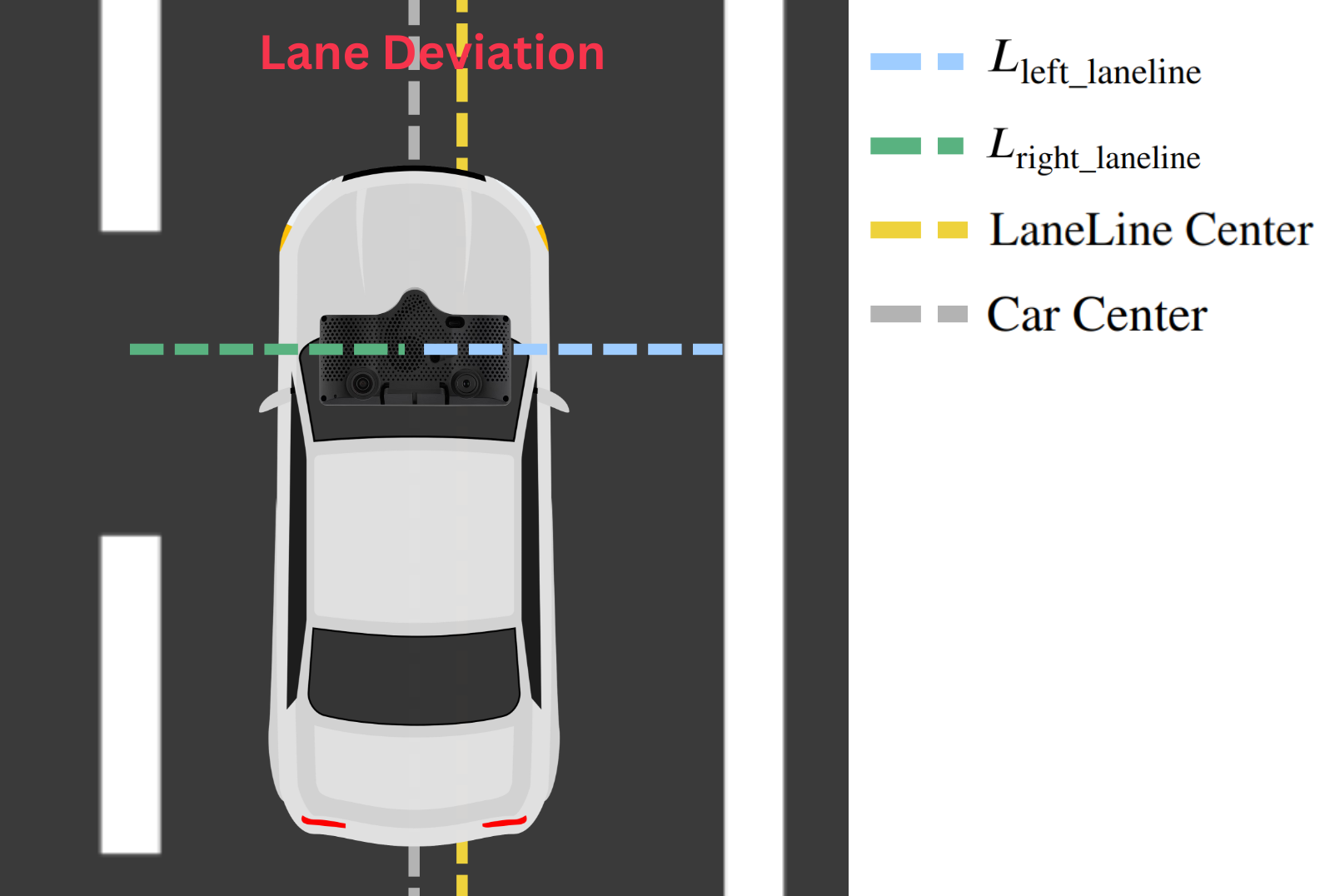}
  \caption{Lane deviation calculation}
  \vspace{-1.0em}  
  \label{fig:formula_exp1}
\end{wrapfigure}

We present the methodology for calculating the LKA error, also known as the \emph{lane centering error} or \emph{lane deviation}, which serves as a metric to evaluate the performance of LKA systems in maintaining the vehicle's position at the center of the lane.

As discussed, the Comma Three-X device was centrally mounted on the vehicle's interior windshield, enabling direct use of OpenPilot's computed values for \texttt{left\_laneline} and \texttt{right\_laneline}, which represent the distances from the device's camera to the left and right lane lines, respectively, where positive distances indicate the right. The LKA error is calculated using the formula: \( \text{LKA Error} = (L_{\text{left\_lane}} + L_{\text{right\_lane}}) / 2 \). A positive LKA Error indicates leftward deviation, while a negative value indicates rightward deviation.

\begin{wrapfigure}{l}{0.47\textwidth}
  \vspace{-0.4em} 
  \centering
  \includegraphics[width=\linewidth]{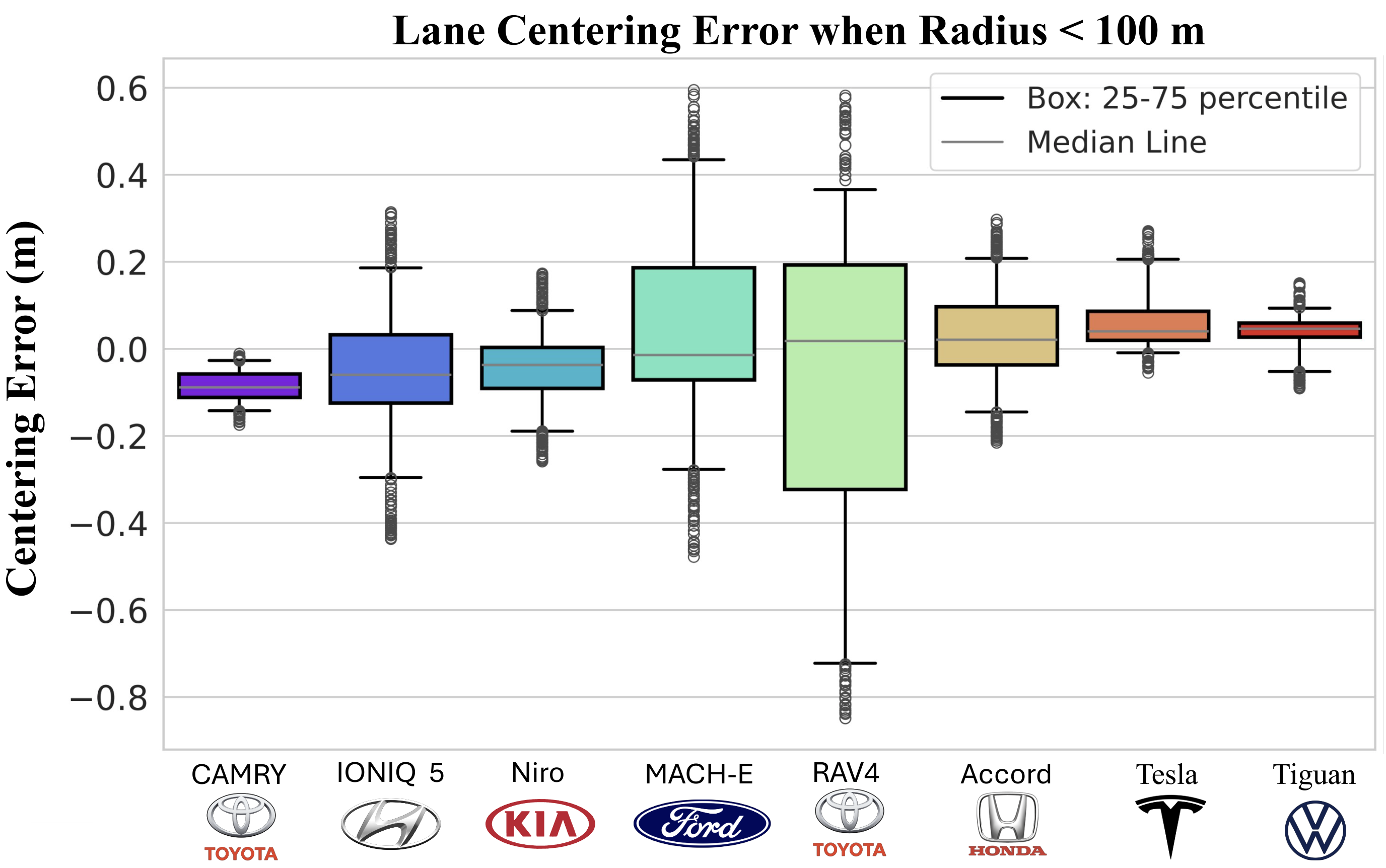}
  \caption{LKA error under normal conditions}
  \label{fig:lka_normal}
  \vspace{-0.4em} 
\end{wrapfigure}

We analyzed the LKA error for each vehicle model under normal driving conditions with road curvature less than 0.05. The results, illustrated in Fig.\ref{fig:lka_normal}, indicate that all vehicles maintain an LKA error around 0.15 meters, with the majority exhibiting errors less than 0.1 meters. This suggests that the LKA systems are effective in keeping the vehicle centered within the lane under typical conditions.

Although the LKA system generally meets expectations under normal driving conditions, real-world traffic scenarios are constantly evolving, and vehicles must adapt to a wide range of environments. In many of these more challenging conditions, the LKA’s performance differs significantly from what we see in standard situations. In the following section, we will examine common factors leading to LKA failures found in the OpenLKA dataset. We will delve into these issues from the perspectives of perception, decision-making and planning, as well as control in autonomous driving systems.

\subsection{LKA Characteristics on Curves}
\label{subsec:curves}

\begin{figure}[htbp]
  \centering
  \includegraphics[width=\textwidth]{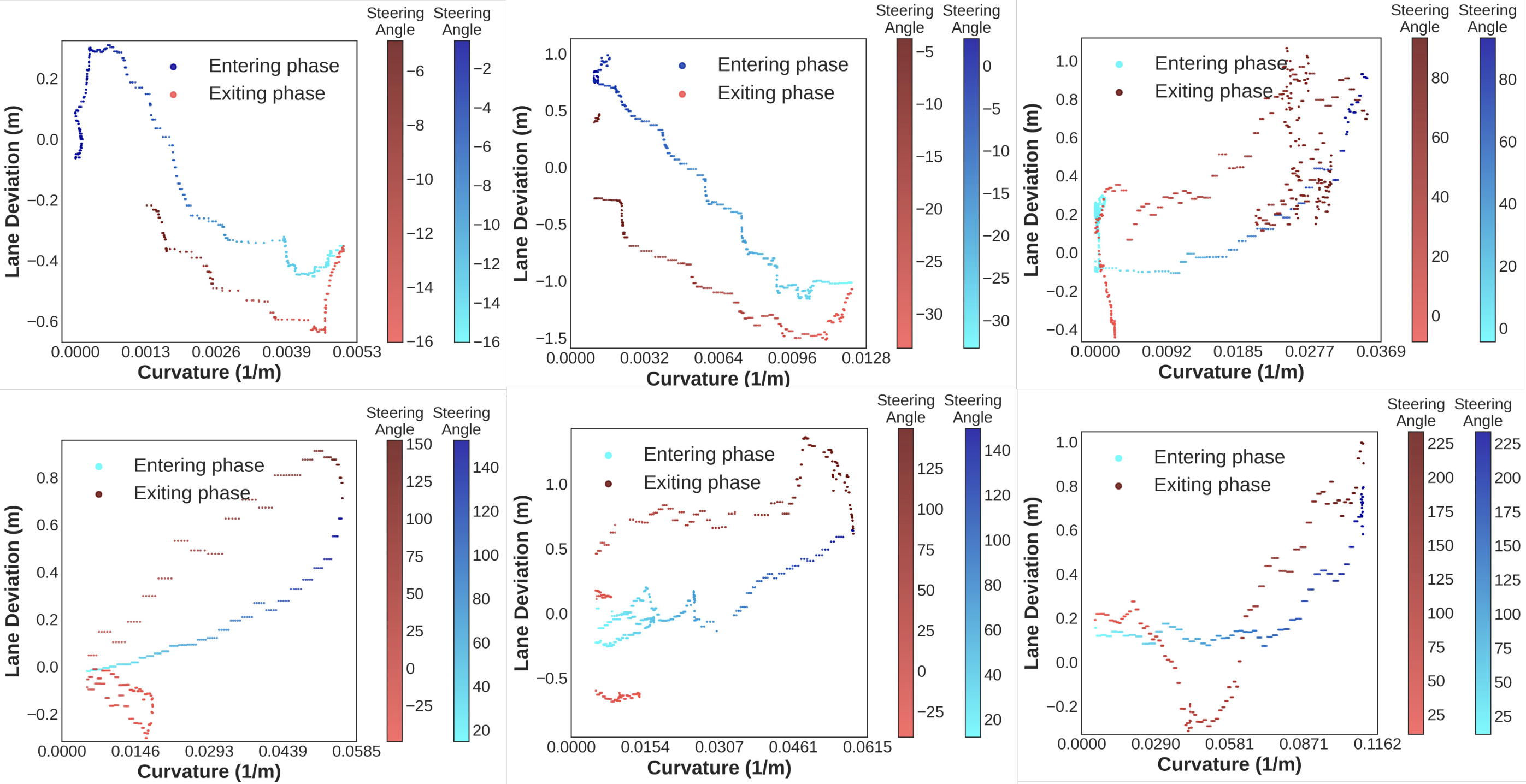}
  \caption{Relationship between Lane deviation and Curvature}
  \label{fig:LDC}
\end{figure}

\begin{wrapfigure}{r}{0.4\textwidth}
  \vspace{-0.5em} 
  \centering
  \includegraphics[width=\linewidth]{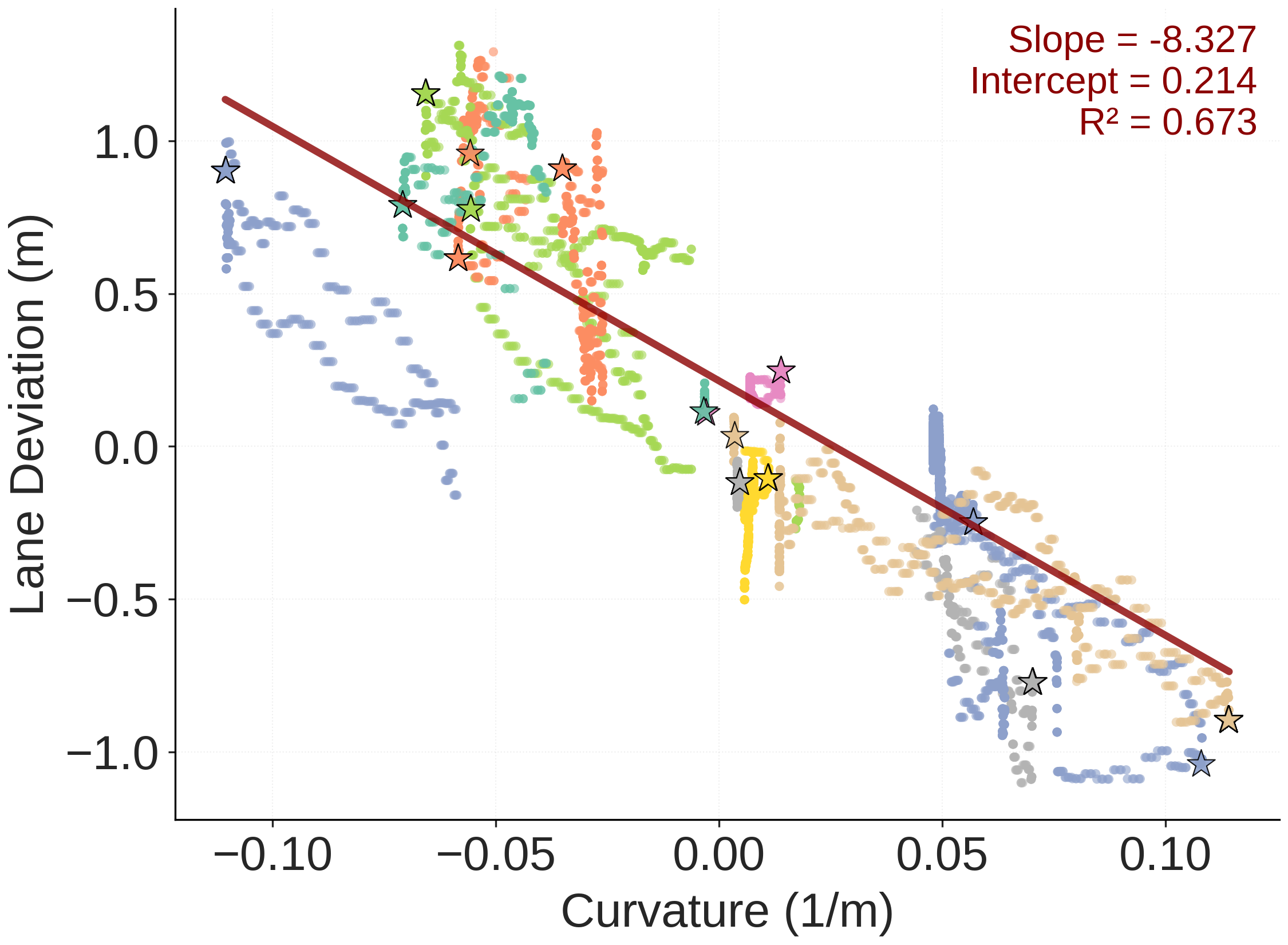}
  \caption{Lane deviation in different curvature}
  \vspace{-0.5em} 
  \label{fig:LDCFIT}
\end{wrapfigure}

Although the LKA system performs well on normal road sections, we found that the LKA system often has large lane deviation during our tests. The most intuitive feeling is concentrated on the road sections with large curvature, where the LKA system often deviates from the center of the lane. As shown in Fig~\ref{fig:LDC}, we analyze the scatter plots of Curvature-Lane deviation for different curvatures of the roadway entering the curve and leaving the curve gradually. These six figures demonstrate that the LKA system tends to have larger lane deviations on road sections with high curvature. Initially, scatter plots of curvature versus lane deviation reveal a pronounced hysteresis effect: the lane deviation for the same curvature differs between entering and exiting phases of the curve. This phenomenon suggests that the vehicle’s lateral position is path-dependent and closely related to road curvature. Furthermore, as curvature increases, lane deviation tends to become larger, indicating the LKA system may struggle to maintain a centered lane position on sharper curves.

We select the driving data of different vehicle models under the scenarios of large curvature in several identical test sections in the data set. Considering that entering and exiting a curve is a process of increasing and decreasing actual curvature, respectively, we preprocessed these data to select the ones with the absolute value of actual curvature in the first 60\%, which means that we selected the driving data that are completely on the curve. Then, we plot a scatter plot of curvature-lane deviation for these data, and select the points where the absolute value of curvature is in the top 2\% and lane deviation is in the top 5\% as the star points, which are used to mark the Apex region of the curve, as shown in Fig.~\ref{fig:LDCFIT}. We regressed all the scatter points and obtained a slope of the fitted line of -8.33, meaning that as curvature (in 1/m) increases, lane deviation generally becomes more larger. Interpreted practically, this suggests that on sharper curves, the vehicle consistently shifts its lateral position in the lane.  When the curvature is larger than 0.05, the lane deviation will come to almost 1.5m, causing serious safty problems issues.

\subsection{LKA Limitations in Challenging Scenarios}
\label{subsec:limitations}

\subsubsection{LKA Limitations in Perception}
\label{subsubsec:perception_limits}

In the realm of autonomous driving, perception encompasses the system’s ability to sense, interpret, and understand its immediate surroundings. This typically involves processing raw sensor data—collected by cameras, lidars, radars, and other sensors—to accurately identify lane lines, detect obstacles, and gauge the vehicle’s position relative to road boundaries. Robust perception is fundamental to safe and reliable lateral control. Without a clear and consistent perception of the driving environment, even advanced LKA systems struggle to maintain stable and accurate lane centering, especially under challenging conditions.

There can exist multiple contributing factors that lead to poor lane line perception in OpenLKA:

\textbf{Contrast between pavement and lane markings}

OpenLKA encompasses a broad spectrum of pavement and lane marking color combinations. Our experimental results indicate that certain pairings significantly degrade LKA performance, frequently leading to system disengagement. Table~\ref{tab:LKA_Impact} summarizes the effects of various pavement and lane marking colors on LKA stability. Notably, conditions such as white pavement with white lane lines often induces excessive steering corrections, resulting in suboptimal handling and, in some instances, lane departures that pose serious safety risks. It is worth mentioning that we have found that sometimes white pavement with black lane lines or pavement with white markings bordered by black can cause deviations in LKA, which is consistent with a number of previous assessments  \cite{pike2024assessing}. 

\begin{figure}[htbp]
  \centering
  \includegraphics[width=0.8\textwidth]{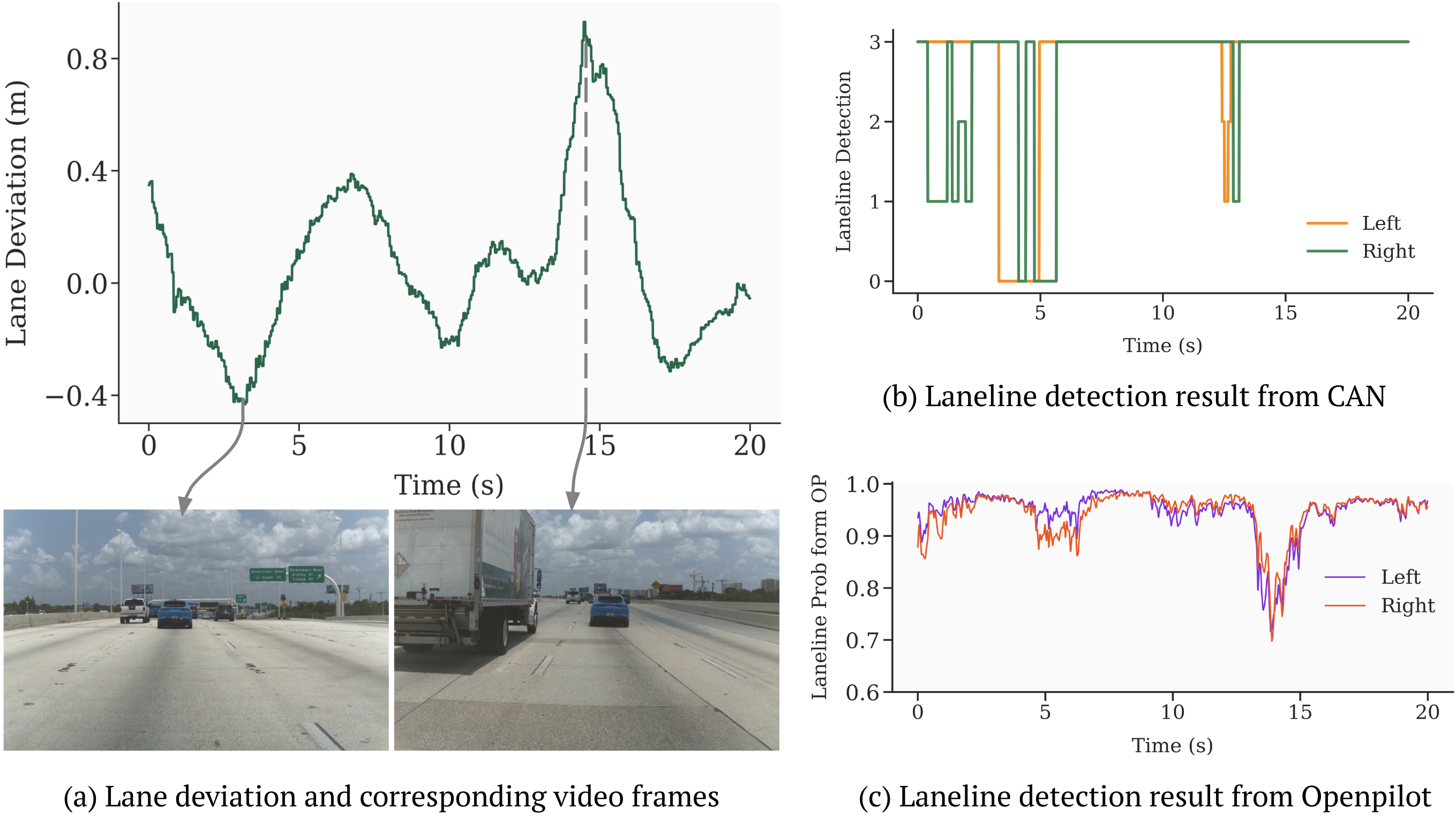}
  \caption{LKA Performance in Unusual Pavement Conditions}
  \label{fig:laneline_contrast}
\end{figure}

Figure~\ref{fig:laneline_contrast}(a) illustrates a particularly hazardous scenario under white pavement with white lane lines, where the vehicle veers left and nearly exits its designated lane, with lateral deviations exceeding 0.8 meters. The presence of a large truck passing on the left further intensifies the situation, necessitating prompt human intervention to avert a potential accident. Inadequate or incorrect intervention can lead to critical safety hazards.

Figures~\ref{fig:laneline_contrast}(b) and (c) present the corresponding lane line detection results from both the vehicle’s system and openpilot. As shown in Figure~\ref{fig:laneline_contrast}(b), lane detection probabilities for both left and right lane lines fluctuate considerably, oscillating among non-detection, faint detection, and lane departure indications. These instability patterns severely undermine driving safety. Even openpilot, which employs state-of-the-art lane detection algorithms, exhibits notably low detection probabilities for both left and right lane lines under these conditions.

\captionsetup[table]{labelfont=bf, font=small}
\begin{longtable}{p{3.5cm}p{3.5cm}p{7.5cm}}
\caption{Effects of Different Road Surface and Lane Marking Colors on LKA Performance}
\label{tab:LKA_Impact} \\

\toprule
\textbf{Road Surface Color} & \textbf{Lane Marking Color} & \textbf{LKA Impact Description} \\ 
\midrule
\endfirsthead

\toprule
\textbf{Road Surface Color} & \textbf{Lane Marking Color} & \textbf{LKA Impact Description} \\ 
\midrule
\endhead

\midrule
\multicolumn{3}{r}{\textit{Continued on next page}} \\
\endfoot

\bottomrule
\endlastfoot
Black & White & High contrast, good detection \\
Black & Yellow & High contrast, good detection \\
White & White & Low contrast, potentially difficult detection \\
White & Black & High contrast, but detection stability may vary* \\
White & Yellow & Moderate contrast, fair detection \\
White & Black and White & Moderate contrast, detection stability may vary \\
\end{longtable}

\textbf{Perceived failure due to changes in environment and traffic scenarios}

\begin{figure}[htbp]
  \centering
  \includegraphics[width=\textwidth]{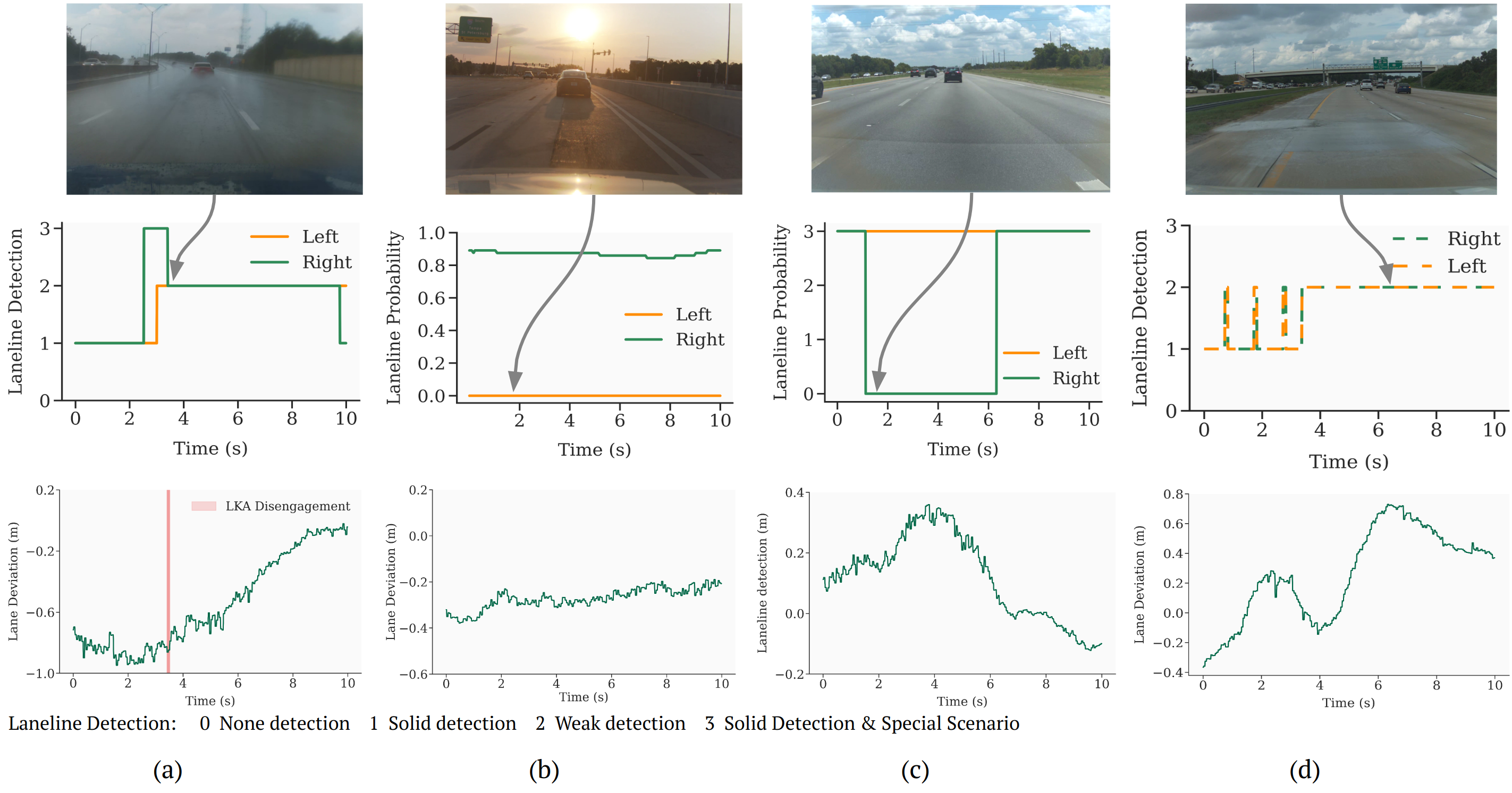}
  \caption{Analysis of LKA Performance in different perception scenarios}
  \label{fig:perception_cases}
\end{figure}

OpenLKA contains various extreme weather conditions characteristic of Florida, including sudden heavy rain and intense sunlight glaring. Testing vehicles under these extreme environments not only enriches the diversity of our dataset but also provides critical data for assessing the authenticity and robustness of LKA performance. Figure~\ref{fig:perception_cases}(a) illustrates a common scenario within our dataset where a sudden heavy downpour destabilizes LKA performance. Specifically, heavy rain and road surface water significantly impair lane line detection, the detection result is very weak, according leading to a lane deviation over 1.0 meter to the right, causing the vehicle to struggle in maintaining lane centering. The LKA system directly disengages at around 3.5 seconds, causing the driver to slow down to avoid potential hazards in blurred vision.

Beyond the heavy rain conditions shown, intense sunlight glaring poses substantial challenges by creating strong reflections on the pavement and lane markings, which adversely affect road detection and, consequently, LKA performance. These reflections can lead to inaccurate lane detections and erratic steering behaviors, undermining the LKA system's ability to maintain the vehicle's position within the lane. This underscores the necessity for vehicle manufacturers to enhance the robustness of their vision systems, radar, and lidar technologies to reliably detect lane markings under diverse and deteriorating environmental conditions. Strengthening these detection capabilities is essential to ensure the consistent and safe operation of LKA systems across a wide spectrum of real-world driving scenarios.

Figure~\ref{fig:perception_cases}(c) shows the problems that occur in LKA in the case of a lane change. The disappearance of the traffic lane lines and the widening of the road cause the LKA to be unable to detect the right-hand marking, thus causing the vehicle to deviate abruptly to the right. In this perceptual scenario, the algorithm should itself be able to predict a reference in the absence of a road marking or plan along a detectable marking. In addition, the algorithm should be able to learn the behavior of the vehicle in front of it and plan when there is a vehicle in front of it. Figure~\ref{fig:perception_cases}(d) illustrates a case where the edge lines are not detected because the road markings are worn out. In this case, the left and right edges are not detected due to wear and tear, and the lane deviation reaches 0.8 at one point. 

In the above case, due to the changes in the external environment and the traffic environment, one of the important references of LKA, Laneline, cannot be detected correctly, which leads to a lot of problems in LKA. However, for human drivers, when the above problems occur, they can still refer to a relatively clear road marking, refer to a certain safe driving pattern or follow the driving trajectory of the surrounding vehicles to plan a reasonable route, and still be able to keep the vehicle in the center of the lane, without sudden change of direction or large lateral acceleration. Therefore, the performance of LKA in the above scenario is problematic, and there is still a big gap between safe and comfortable autonomous driving.

\subsubsection{LKA Limitations in Planning}
\label{subsubsec:planning_limits}

\textbf{Lane Line Transitions}

\begin{figure}[htbp]
  \centering
  \includegraphics[width=\textwidth]{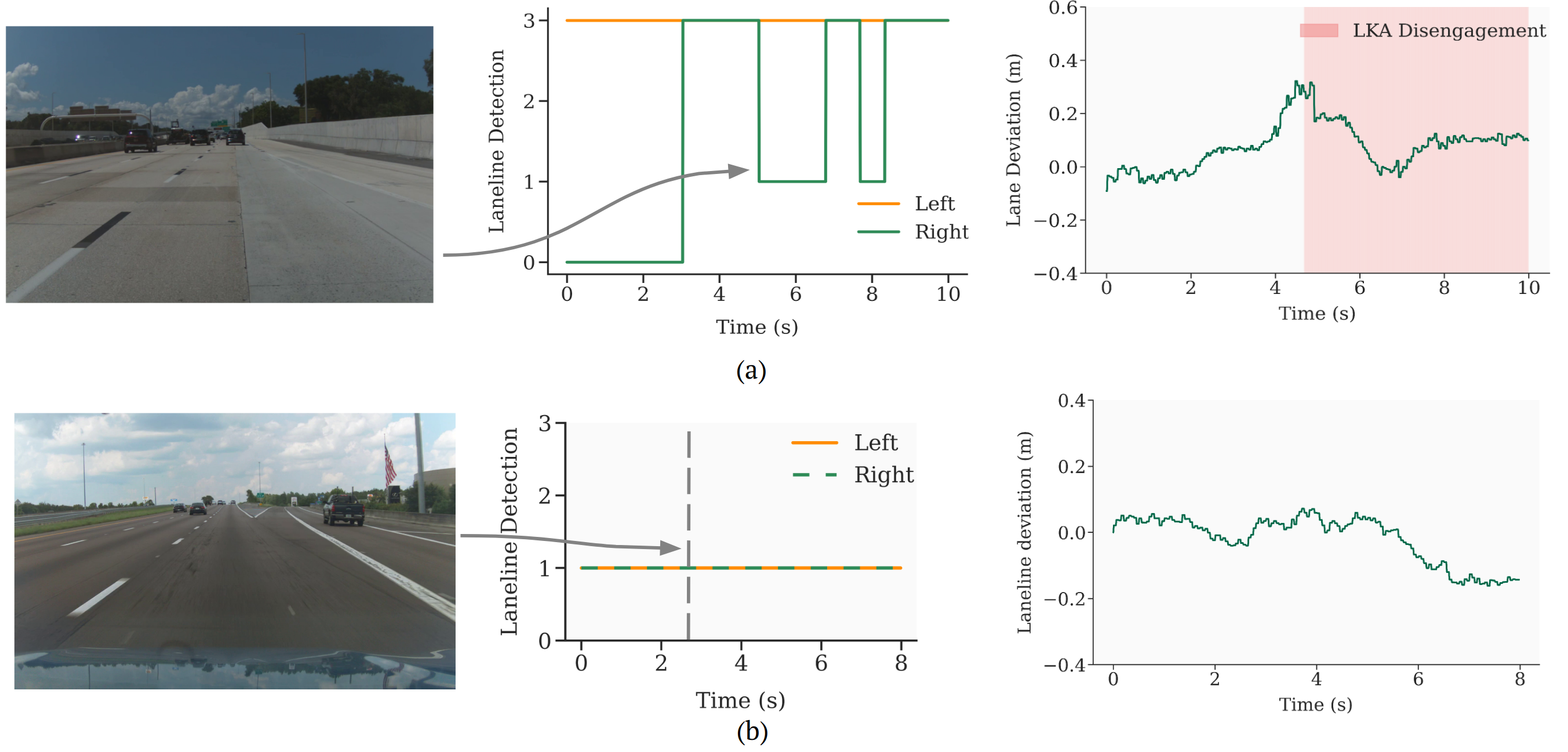}
  \caption{LKA Performance During Lane Line Transitions}
  \label{fig:lanechange}
\end{figure}

Within OpenLKA dataset, scenarios where lane line transitions lead to LKA disengagements are frequently observed. Lane line transitions typically occur due to variations in road width or structural modifications, such as lane narrowing, lane merging, or discontinuities in lane markings. Figure~\ref{fig:lanechange}(a) illustrates a situation in OpenLKA dataset where the vehicle's right lane line bifurcates into a main road and a side road. During this period, as shown in Figure~\ref{fig:lanechange}(b), the detection probability of the vehicle's right lane line drops below 0.2, preventing the vehicle from maintaining its position within the current lane line and causing it to veer towards the right side of the road. Without timely human intervention, the vehicle would likely merge onto an off-ramp, posing significant safety hazards. Similar scenarios are prevalent in our dataset, highlighting the commonality of such conditions and the substantial risks they introduce to the effective operation of LKA systems.

Figure~\ref{fig:lanechange}(a) illustrates a situation in OpenLKA dataset where the vehicle's right lane line merged from two roads into a main road. During this period, though the detection probability of the vehicle's left and right lane line is stable, the vehicle still can't maintain its position within the current lane line and causing it to veer towards the right side of the road. This is because the vehicle's sensoring cmera detect the wrong line, causing the lane width have a sudden change. At the same time, the LKA system correct the vehicle to run in the center of the lane. If a vehicle is accelerating right behind, without timely human intervention, the vehicle would likely merge onto the middle of the off-ramp and current lane, posing significant safety hazards. Similar scenarios are prevalent in OpenLKA dataset, shown in Figure~\ref{fig:lanechange}(b). However, some vehicles have more flexible planning methods. We found that Figure~\ref{fig:lanechange}(b) can make good planning based on the changes in the road line, thus avoiding driving in the middle of the two forked roads. In addition, not only when the perception is clear, but also when there is a problem with the perception part, the planning part should design appropriate algorithms to ensure the safety of the vehicle, thereby improving the robustness of the autonomous driving system.

\subsubsection{LKA Limitations in Control}
\label{subsubsec:control_limits}

\begin{wrapfigure}{r}{0.4\textwidth}
  \centering
  \includegraphics[width=\linewidth]{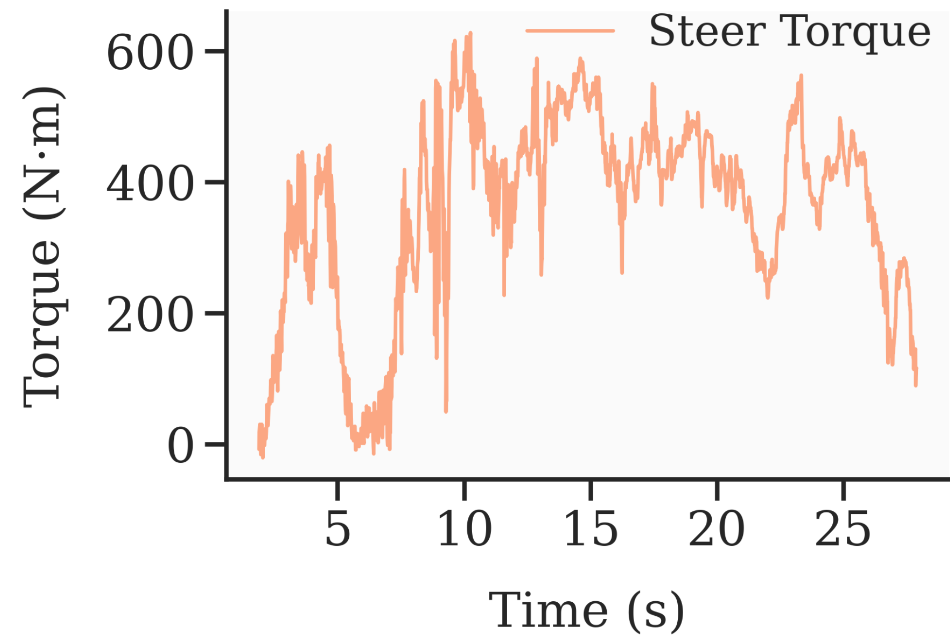}
  \caption{Analysis of Steering Torque Variations}
  \label{fig:control_torque}
\end{wrapfigure}

The performance of LKA systems on high-curvature roads is closely tied to the lateral acceleration required to maintain the desired path. This lateral acceleration is not only influenced by the road geometry but also by the vehicle's dynamic characteristics, such as roll-induced forces. To account for these factors, the compensated lateral acceleration can be expressed as:

\begin{equation}
    a_{\text{lateral}} = \kappa \cdot v^2 - (\text{roll} \cdot g)
    \label{eq:lateral_acceleration}
\end{equation}

where \(a_{\text{lateral}}\) represents the lateral acceleration (m/s²), \(\kappa\) is the path curvature (1/m), \(v\) is the vehicle speed (m/s), \(\text{roll}\) is the roll angle (rad), and \(g\) is the gravitational acceleration (9.81 m/s²).

Given the critical role of lateral acceleration in determining vehicle stability and trajectory tracking, LKA systems employ advanced steering control strategies to maintain the vehicle's path. These systems primarily rely on two steering control methods: \emph{steering angle control} and \emph{steering torque control} \cite{PARK2021103213,na2020torque}. 

Steering angle control directly specifies the desired steering wheel angle, guiding the vehicle along a predefined trajectory. However, this approach has inherent limitations. It is constrained by the mechanical limits of the steering system, which can lead to abrupt steering inputs if not carefully managed, potentially impacting vehicle stability and passenger comfort \cite{hegazy2009vehicle, zhou2021vehicle}. Additionally, factors such as actuator saturation and limited bandwidth reduce its effectiveness in scenarios with rapidly changing driving conditions \cite{wang2014rolling}.

\begin{figure}[htbp]
  \centering
  \includegraphics[width=\textwidth]{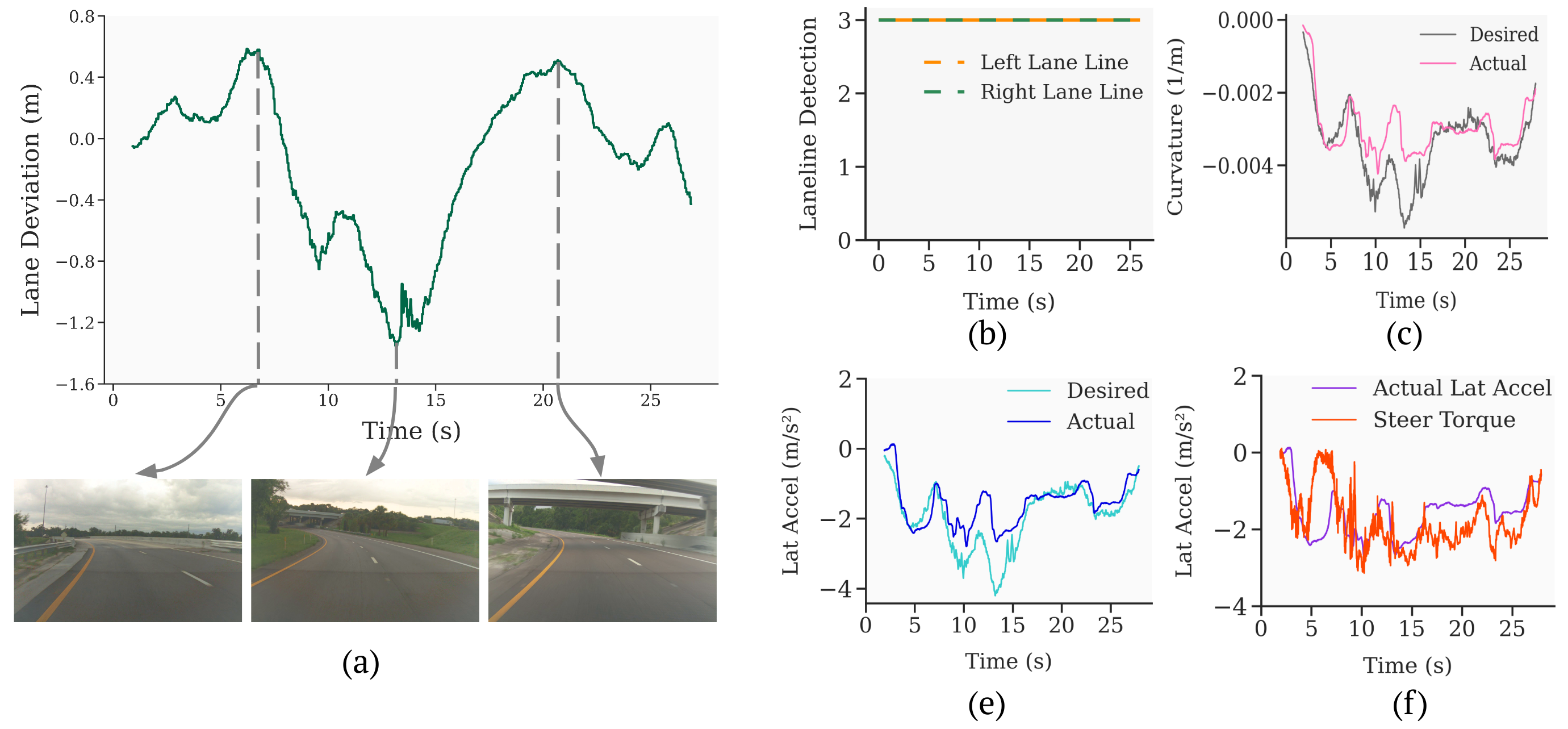}
  \caption{Impact of Sharp Curves on LKA Performance}
  \label{fig:lka_disengagement_on_curve}
\end{figure}

In contrast, steering torque control utilizes the vehicle's \emph{Electric Power Steering} (EPS) system to apply torque, assisting the driver in steering maneuvers. While this method is less intrusive and more intuitive for the driver, it faces its own set of challenges. Torque control may lack the authority to execute significant steering corrections required on sharp curves, particularly when torque limits are imposed \cite{park2021research}. In high-speed scenarios or situations necessitating rapid steering adjustments, the system may struggle to generate sufficient torque to promptly achieve the desired steering angle, as it is designed to assist rather than override driver input \cite{dixon2023exploring}.

Furthermore, many manufacturers, such as Volkswagen, implement \emph{torque limits} in their control systems to ensure safety during assisted steering. These torque limitations prevent sudden or excessive steering actions that could startle the driver or compromise vehicle handling \cite{Volkswagen2020}. However, they can also restrict the LKA system's ability to perform necessary steering corrections, particularly on tight curves or during emergency maneuvers \cite{badue2021self}. This trade-off highlights the delicate balance between safety, driver comfort, and system performance, which remains a central challenge for LKA systems on complex road geometries.

As a result, when navigating high-curvature road segments, LKA systems may struggle to maintain the vehicle's position within the lane center. Fig.~\ref{fig:lka_disengagement_on_curve} illustrates instances of degraded LKA performance on such roads, showing increased lane-centering errors and occurrences of lane departure. Fig.~\ref{fig:lka_disengagement_on_curve}(a) display the scenarios happens on sharp curve, where the lane deviation is over 1.2 meters during the turn. Fig.~\ref{fig:lka_disengagement_on_curve}(c) shows the curvature and (e) shows the Desired\&Actual Lateral acceleration. We can see from it that the control algorithm still face some challenges in large curvature. The combination of torque limitations and the inherent constraints of steering control methods impedes the system from applying adequate steering inputs to follow sharp curves accurately, leading to potential safety concerns and highlighting a significant issue in current LKA technology.

To address this challenge, it is imperative for vehicle manufacturers to optimize the balance between safety constraints and the operational requirements of LKA systems. Potential solutions include adaptive torque limit strategies that adjust based on road geometry \cite{chen2019extracting} and driving conditions, or advancements in control algorithms that enhance the system's ability to handle sharp curves without compromising safety \cite{elvik2019more}. Enhancements in steering control methods \cite{li2021enhancement}, such as developing more robust actuator designs \cite{fogla2023vehicle, swain2021neural} to mitigate issues like actuator saturation and limited bandwidth, could also improve performance in rapidly changing driving conditions.

\subsubsection{Combination of Adverse Conditions}
\label{subsubsec:combined_conditions}

\begin{figure}[htbp]
  \centering
  \includegraphics[width=\textwidth]{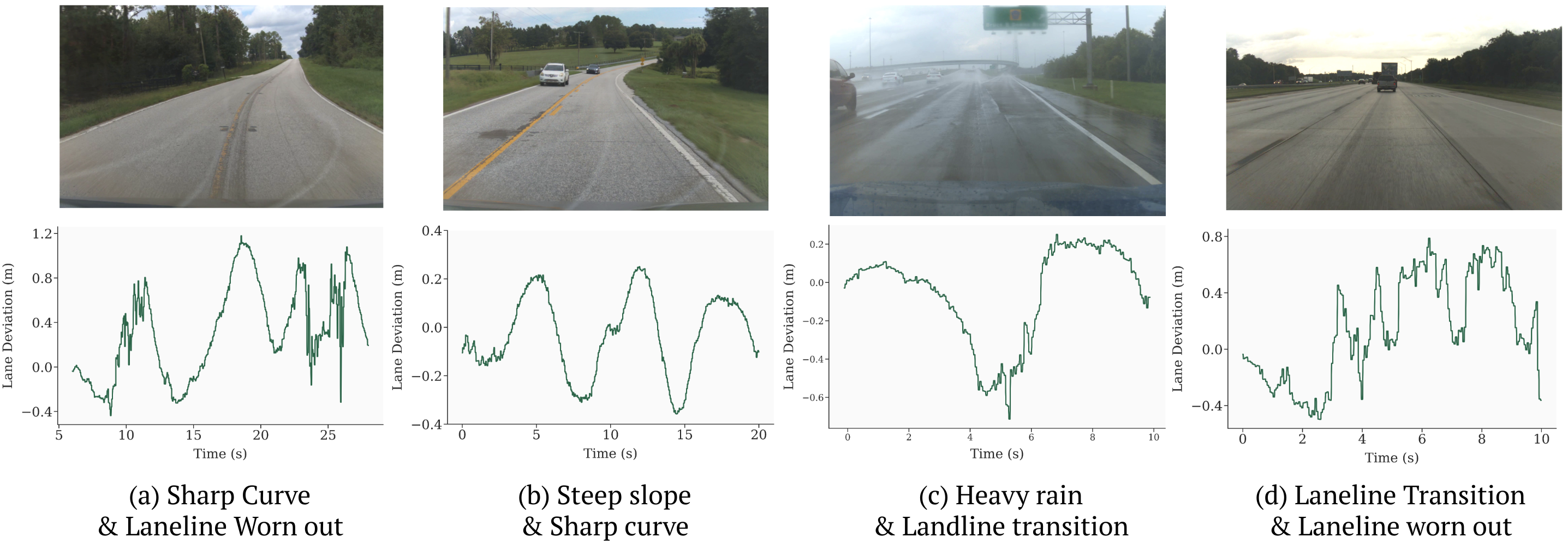}
  \caption{Critical Performance Factors and Their Combinations in LKA Systems}
  \label{fig:comb_cases}
\end{figure}

In the previous section, we listed the main factors that can lead to poor performance of LKA in terms of perception, decision planning, and control. When these factors are superimposed on each other, the performance of LKA becomes more unstable and disengagement becomes more frequent. By analyzing the OpenLKA dataset, we conclude that the performance of LKA becomes worse when the following factors are superimposed, which is shown in Fig.~\ref{fig:comb_cases}.

\textbf{Sharp curve + Laneline wornout} In scenarios involving a combination of sharp curves with worn-out lane markings, LKA often struggles to keep the vehicle within the lane, increasing the risk of crossing lane boundaries. Without timely human intervention, particularly in oncoming traffic situations, such conditions can pose severe safety risks. The OpenLKA dataset captures vehicle driving data under these challenging scenarios, recorded without human intervention on closed sections of road and under the supervision of experienced drivers. As illustrated in Fig. X, the vehicle not only crossed into the oncoming lane but also showed a tendency to veer off the road entirely. In sharp curve scenarios, LKA's ability to maintain lane position is particularly fragile. A single error in lane line detection can cause the system to follow an incorrect lane marking or fail altogether, further exacerbating the danger.

\textbf{Steep slope + Sharp curve} In scenario involving a combination of steep slope with sharp curve, LKA often fails to provide good lateral control. Unreasonable lateral acceleration causes the vehicle to deviate too far from the center of the lane, and may even cause LKA to fail.

\textbf{Heavy rain/Glaring + Lane line contrast/Lane line transitions} In scenarios involving heavy rain or glaring sunlight combined with poor lane line contrast, LKA systems often face significant challenges in accurately detecting lane boundaries. Adverse weather conditions like heavy rain can obscure lane markings, while intense sunlight can create glare that washes out lane line visibility. These factors reduce the contrast between the lane lines and the road surface, making it difficult for LKA cameras and sensors to function effectively. Without reliable lane detection, the system may fail to maintain proper lane positioning, increasing the risk of unintended lane departures. 

\textbf{Laneline transition + Laneline wornout} In scenarios where lane line transitions occur alongside worn-out lane markings, LKA systems often struggle to provide consistent lateral control. Lane line transitions—such as merges, splits, or shifts due to construction—require the LKA system to accurately interpret changing lane configurations. When these transitions are compounded by faded or worn-out lane lines, the system's ability to track the correct path is severely compromised. This increases the likelihood of the vehicle drifting out of its intended lane or making incorrect steering adjustments.

\subsection{Lack of Flexibility: LKA versus Human Control}
\label{subsec:flexibility}

\begin{figure}[htbp]
  \centering
  \includegraphics[width=0.75\textwidth]{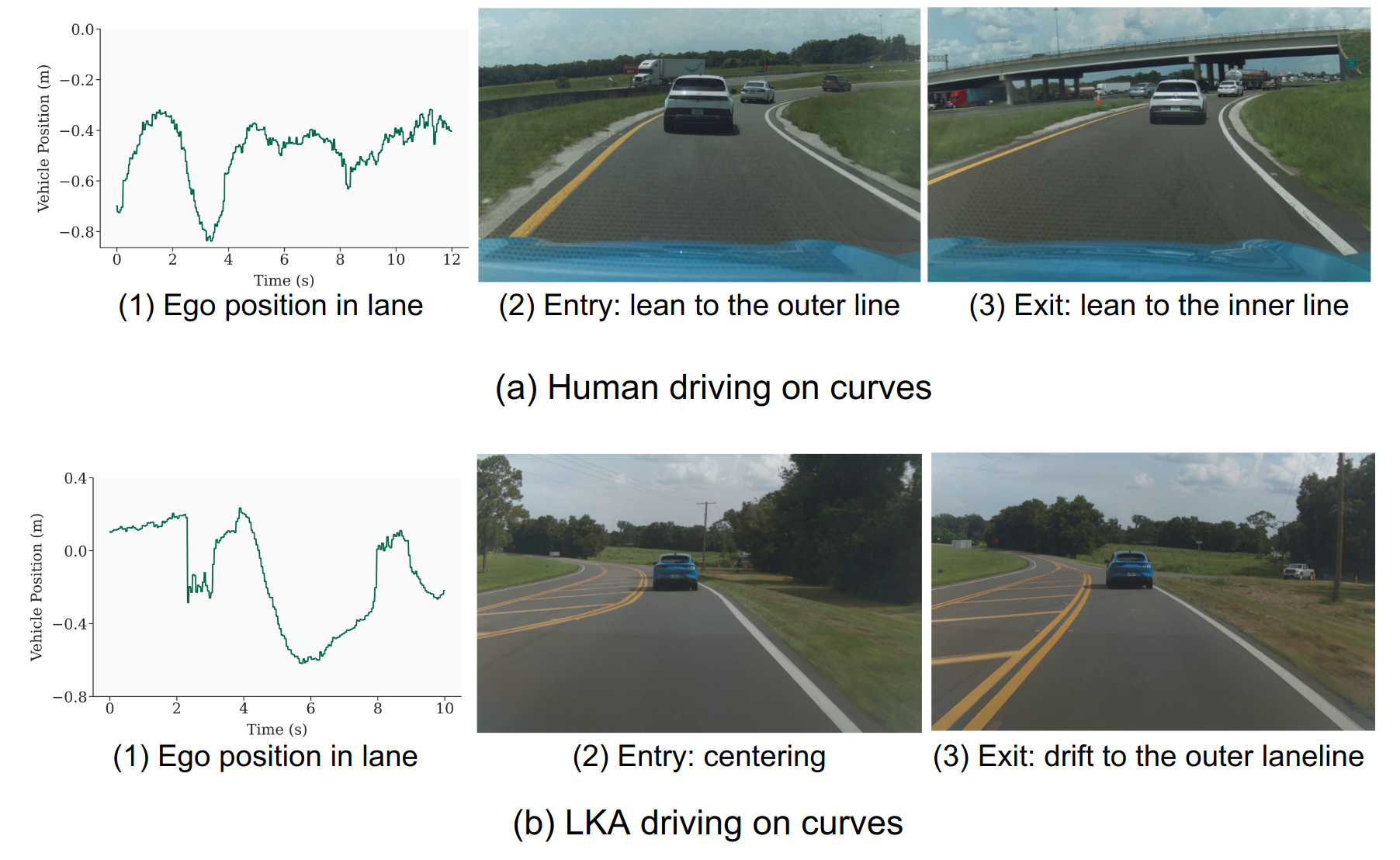}
  \caption{Comparative Analysis of LKA and Human Driving Strategies}
  \label{fig:HVsAV1}
\end{figure}

Experimentally, we found that the LKA system always tries to keep the vehicle in the center of the lane. While LKA systems' rigid adherence to following the center-line of the roadway accomplishes the driving task in most cases, we believe it poses a significant traffic safety hazard. In contrast to the rigid LKA lateral control, a human driver can synthesize and analyze the information of the surrounding environment to make safer decisions. 

\begin{figure}[htbp]
  \centering
  \includegraphics[width=0.9\textwidth]{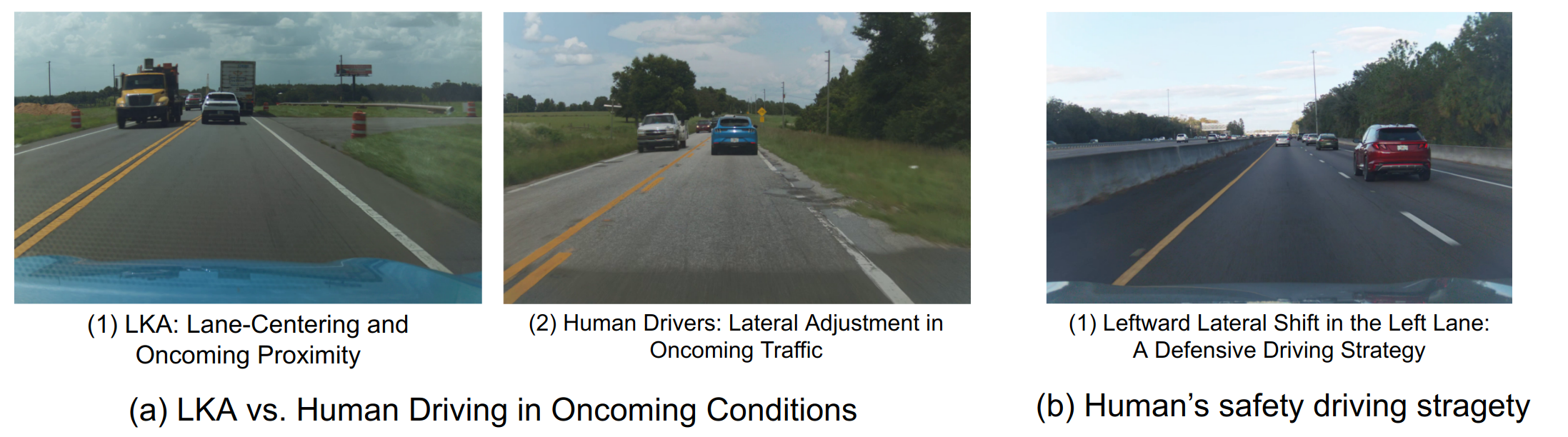}
  \caption{Behavioral Differences Between LKA and Human Drivers}
  \label{fig:HVsAV2}
\end{figure}

For example, in Fig~\ref{fig:HVsAV1}, we compare the difference between LKA and Human Driving in handling curves. As shown in the first half, if the curve is sharp, the human driver will use Corner Apexing to first drive to the opposite side of the curve, and then aim at the inside of the curve to make the turn, so as to ensure that the vehicle stays within the lane line all the way through the curve, and does not go out of the curve. In contrast, in the second half of this section, LKA does not use a similar method, but gradually adjusts the steering angle with the curve, which can easily cause the vehicle to drive out of the lane line, causing it to veer off the road. In addition, avoiding vehicles in on-coming scenarios and actively keeping a distance from lateral vehicles at highway speeds are also very important driving strategies. Fig~\ref{fig:HVsAV2}(a) and (b) show that the human driver actively avoids vehicles in the on-coming scenario, vehicles in other lanes of the highway, and special vehicles, respectively. Therefore, for the inflexible LKA system, we should design a more flexible LKA system that conforms to the human driver's style to ensure driving safety. One of the most suitable methods for designing such systems is to use large-scale visual language models. Considering that large generalized VLMs have a large amount of prior knowledge and are capable of making intelligent decisions under the guidance of appropriate CoT prompting, generalized VLMs have an advantage over other AI models. In an attempt to solve the problem of inflexible decision making in LKA, we propose ``i-LKA'' and give a demonstration example in the next section.

\section{Implications of the Dataset}
\label{sec:implications}

\subsection{Implications for Geometry Design and Speed Limits}
\label{subsec:geometry_design}

\begin{figure}[htbp]
  \centering
  \includegraphics[width=0.7\textwidth]{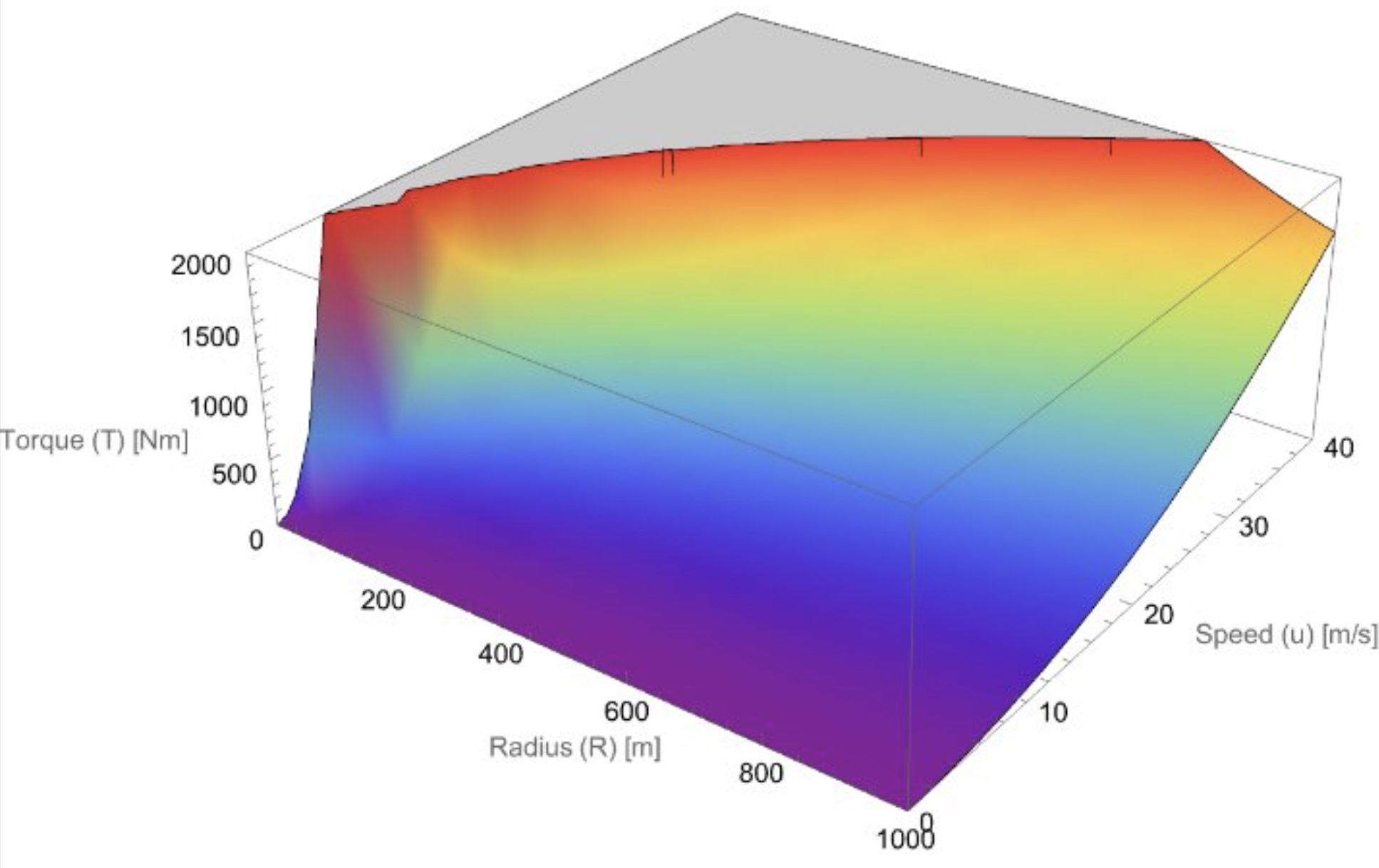}
  \caption{Analysis of Road Geometry Impact on LKA Performance}
  \label{fig:GEO}
\end{figure}

According to basic physics, the lateral acceleration is a multiplication of the curvature of the road, i.e., $1/R$ and the squared speeds. One also needs to consider the road roll as follows:
\begin{equation}
    a_{\text{lat}} = v(x)^2 \cdot \frac{1}{R(x)} - \text{roll}(x) \cdot g
    \label{eq:lateral_acc}
\end{equation}

The steer torque is linearly proportional to the lateral acceleration with a scaling factor $K_a$:
\begin{equation}
    T(x) = K_a \left( \frac{v(x)^2}{R(x)} - \text{roll}(x) \cdot g \right)
    \label{eq:steer_torque}
\end{equation}

To see how rapid the required torque increases when we enter a circular curve, we take the derivative with respect to position $x$ along the direction of travel and get:
\begin{equation}
    \frac{dT}{dx} = K_a \left( \frac{R(x) \cdot 2v(x)\frac{dv(x)}{dx} - v(x)^2\frac{dR(x)}{dx}}{R(x)^2} - g\frac{d(\text{roll}(x))}{dx} \right)
    \label{eq:torque_derivative_x}
\end{equation}

Note that $v = \frac{dx}{dt}$, using the chain rule we have the derivative of the torque with respect to time:
\begin{equation}
    \frac{dT}{dt} = \frac{dT}{dx} \cdot \frac{dx}{dt} = \frac{dT}{dx} \cdot v
    \label{eq:torque_derivative_t}
\end{equation}

Substituting the expression for $\frac{dT}{dx}$ and simplifying, we get:
\begin{align}
    \frac{dT}{dt} &= K_a \left( \frac{R(x) \cdot 2v(x)\frac{dv(x)}{dx} - v(x)^2\frac{dR(x)}{dx}}{R(x)^2} - g\frac{d(\text{roll}(x))}{dx} \right) \cdot v \label{eq:torque_time_1} \\
    &= v \cdot K_a \left( \frac{R(x) \cdot 2a(x) - v(x)^2R'(x)}{R(x)^2} - g \cdot \text{roll}'(x) \right) \label{eq:torque_time_2}
\end{align}

On the transition segment to a circular curve, where the radius $R(x)$ reduces with position, we have:
\begin{equation}
    \frac{dT}{dt} = v \cdot K_a \left( \frac{2a(x)}{R(x)} - \frac{v(x)^2R'(x)}{R^2(x)} - g \cdot \text{roll}'(x) \right)
    \label{eq:torque_time_transition}
\end{equation}

Note that here the radius is decreasing, where $R'(x)<0$. It means that when the transition is more rapid, the required torque rate gets larger. We can also see that deceleration would help reduce the increase of torque.

Let's ignore acceleration and roll, we have some design principle that can be applied to the transition curve before a circular curve with radius $R$:
\begin{align}
    \frac{dT}{dt} &= v(x)^3K_a/R^2(x) \cdot (-R'(x)) \label{eq:design_principle_1} \\
    &= v(x)^3K_a/R^2(x) \cdot \left(-\frac{R_c - \infty}{L}\right) \label{eq:design_principle_2}
\end{align}

\subsection{Implications for Lane Marking Design and Rural Safety}
\label{subsec:lane_marking}

Focusing on the extrinsic causes of LKA failure, we propose a new approach to assess road safety and LKA reliability on this road from the perspective of the transportation sector. Based on the OpenLKA dataset, we use the driving data and the corresponding driving videos in the dataset and the labels obtained by labeling with Prompts and GPT-4o that we designed in the method in Section~\ref{subsec:video_augmentation}, we obtain a prediction model of LKA on road traffic facilities constructed using Random Forest and XGBoost, respectively. 

\begin{figure}[htbp]
  \centering
  \includegraphics[width=1.0\textwidth]{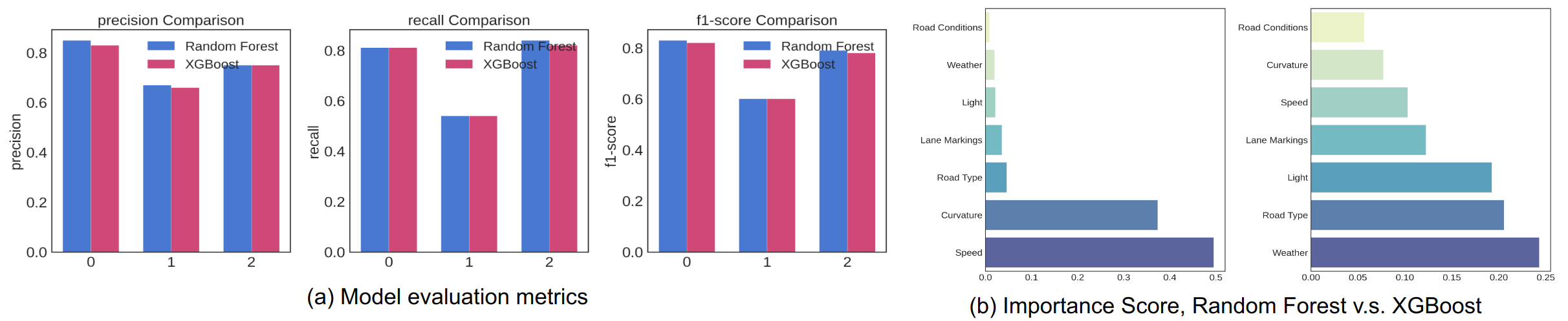}
  \caption{Performance Comparison of Random Forest and XGBoost Models for LKA Prediction}
  \label{fig:model}
\end{figure}

We select a subset of the OpenLKA dataset and choose LKA status and LKA Deviation as output variables. Considering that LKA status is a discrete variable and LKA Deviation is a continuous variable, for the sake of consistency, we categorize LKA in the case of LKA engagement as normal driving and deviation, and categorize LKA in the case of LKA inoperability as LKA disengagement. Specifically, in the case of LKA System Specifically, in the case of LKA System intervention, we select the 75th percentile of LKA Deviation (0.25 m) as the threshold, when LKA deviation is lower than 0.25 m, we consider it as normal driving, and when deviation exceeds 0.25 m, we label it as a serious deviation, and in the case of LKA disengagement, it means that the LKA is not able to be activated normally, which indicates that the environment is very difficult for the LKA to work. For LKA disinterment, it means that LKA cannot be activated normally, which indicates that the environment is very unsuitable for LKA. In the prediction model, we pay more attention to the performance of LKA in deviation and disengagement. For the input variables, we use vehicle speed and road curvature collected directly from OpenLKA, and select time-aligned Road Conditions, Weather, Lighting, RoadType, and Lane Markings from the labels obtained by our annotation method.

\begin{figure}[htbp]
  \centering
  \includegraphics[width=\textwidth]{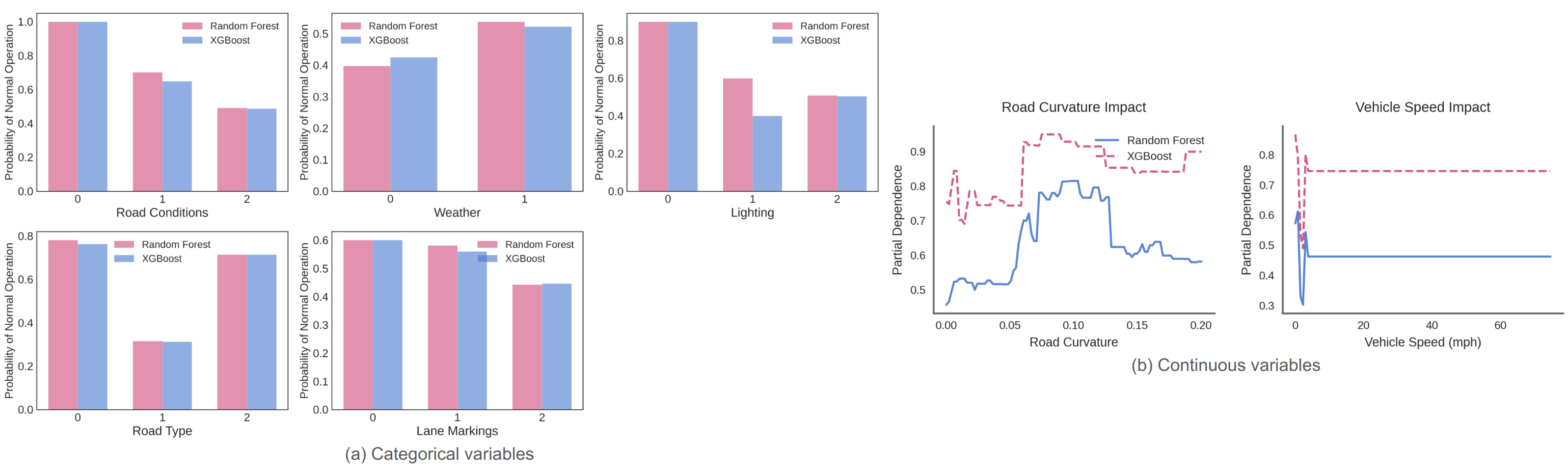}
  \caption{Impact Analysis of Categorical Variables on LKA Performance}
  \label{fig:VImpact}
\end{figure}

We use Random Forest and XGBoost to make predictions, and the prediction results are shown in Figure~\ref{fig:model}. Both models have good performance in terms of prediction accuracy and recall, and we also obtain the effects of different variables on the performance of LKA. We find that speed and curvature are the most important variables that affect the performance of LKA and are most likely to cause LKA problems. In addition, road markings, road lighting and road type and weather are also important. All of these affect the vehicle perception and control to a greater or lesser extent, leading to LKA failure.  Figure~\ref{fig:VImpact} also shows how different variables affect the model. For example, based on our data, we get the judgment that the LKA deviation warning should be set at the place where the road curvature is greater than 0.006, and the speed limit of the vehicle in LKA state should be set, and the recommended speed of the vehicle should be no higher than 60.7mph.

Through our design of this road detection pipeline, the Department of Transportation can easily collect data, expand the data, and analyze the data on the corresponding roads to get how different roads should be regulated to better apply to the current emerging driving technologies. In addition, we have provided the Department of Transportation with a checklist based on the dataset and the results of the analysis, which can be found in the Appendix. By using this checklist, it is possible to conduct faster analysis when designing or evaluating roadways to improve roadway safety.

At the same time, rural roads often present unique challenges that undermine the performance of LKA systems, as highlighted by our dataset. Unlike urban roads, rural roadways frequently suffer from inconsistent maintenance, including faded or absent lane markings, uneven pavement surfaces, and limited use of reflective materials. These issues are further exacerbated by environmental factors such as low lighting conditions, dense foliage obscuring signage, and the prevalence of unpaved shoulders. The disparity in infrastructure quality between urban and rural areas highlights a critical issue of technological inequity. LKA and other ADAS depend on well-defined, high-contrast lane markings to perform effectively. This reliance places rural regions at a significant disadvantage, potentially excluding them from the benefits of these emerging technologies.

\subsection{Implications for AI-assisted LKA Technology Developments}
\label{subsec:ai_implications}

In this section, We will show several uses of the OpenLKA-based dataset for training better self-driving AI models. We will describe LKA from the perspectives of Planning, Perception, Control and Early Warning respectively.

\textbf{LLM-based Human-like LKA Planner}

\begin{wrapfigure}{r}{0.4\textwidth}
  \centering
  \includegraphics[width=\linewidth]{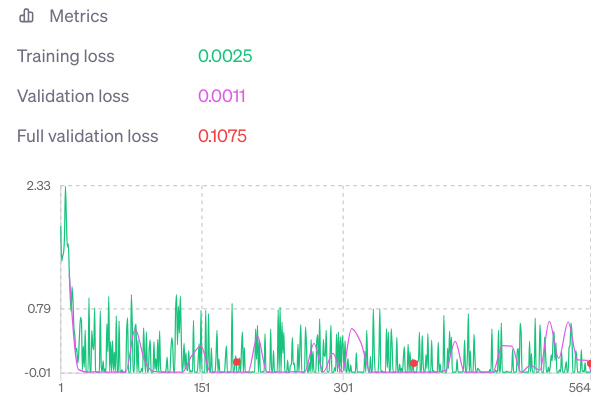}
  \caption{Training and Validation Loss Curves for Model Fine-tuning}
  \label{fig:loss}
\end{wrapfigure}

We introduce a novel OpenLKA-based framework, termed “iLKA,” designed to facilitate the generation of rational, safe, and efficient LKA decisions. This framework combines robust control theory with data-driven approaches, thereby offering a scalable and reliable pipeline for vehicular autonomy. In particular, we present an illustrative case study involving a fine-tuned GPT-4 model enhanced with Chain-of-Thought (CoT) prompt engineering techniques. By processing both CAN messages and front-view imagery, this language model is capable of deriving appropriate actions and providing nuanced control outputs based on high-level situational understanding. Furthermore, to promote the advancement of LKA research, we propose that our insights be translated into language model APIs, thereby enabling broader scenario generation and more comprehensive testing in real-world or simulated environments. In light of the limitations identified in current LKA systems—such as their sensitivity to ambiguous lane markings and complex roadway structures—we advocate constructing an iterative chain-of-thought process. This process would incorporate continuous feedback, error analysis, and refinement to address these limitations, ultimately contributing to the ongoing evolution of state-of-the-art LKA methodologies.

Since each person's driving scenario is different, we propose the 'iLKA' framework to fine-tune the LKA driving style for each person based on his/her own driving data to get a more similar style of driving to human driving, and to enhance his/her flexibility. iLKA accepts the aligned vehicle front vision information and vehicle control information, analyzes them to make the correct LKA decision, and outputs the next control information of the vehicle: steering angle.

\begin{figure}[htbp]
  \centering
  \includegraphics[width=0.8\textwidth]{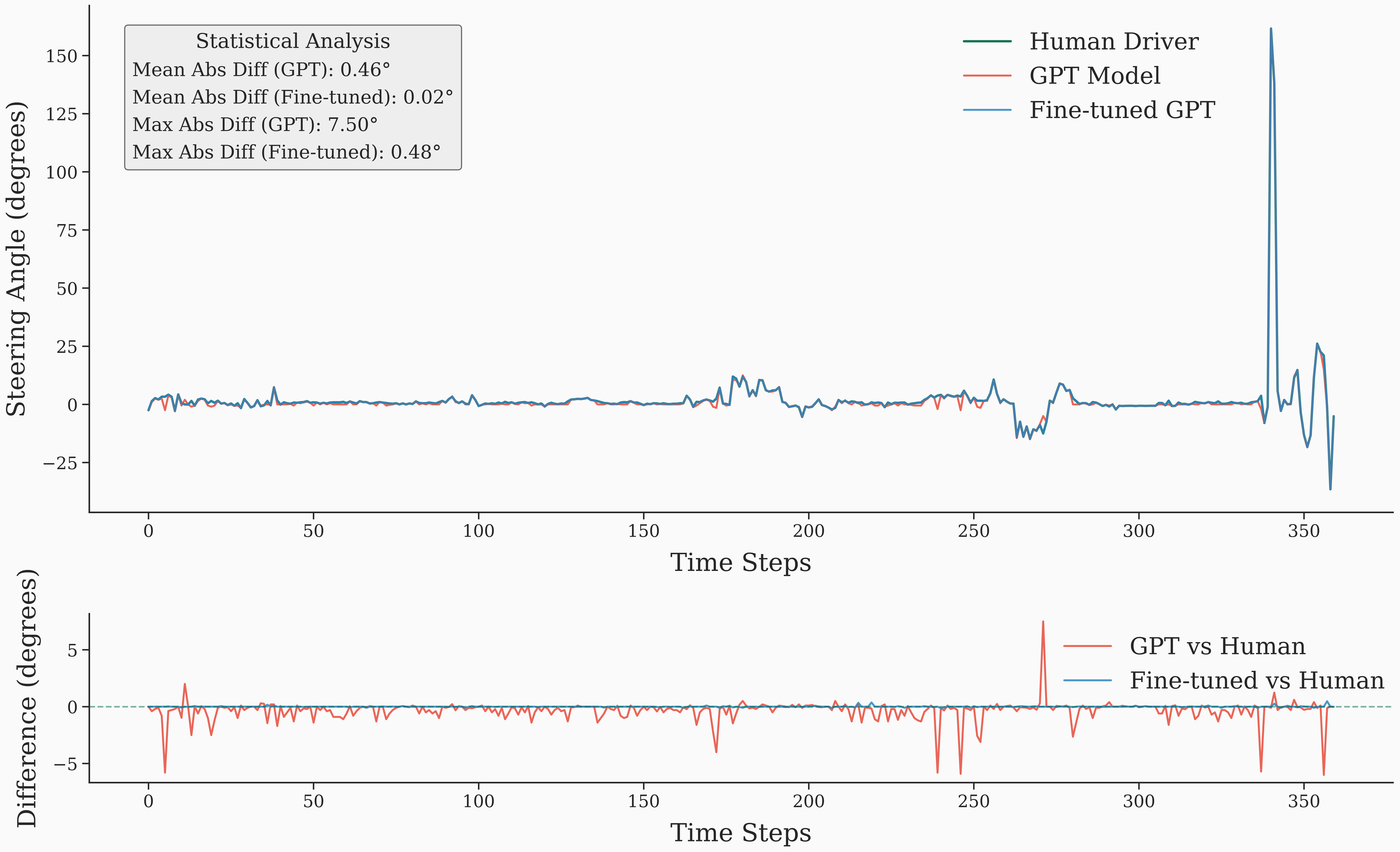}
  \caption{Comparative Analysis of Original and Fine-tuned GPT Models}
  \label{fig:OriginVSFineTuned}
\end{figure}

Specifically, we select an entire segment of driving from the Human dataset in OpenLKA, and we select a segment that ensures that the driver is experienced and has excellent safety awareness, and that his or her lateralized driving is flexible and safe as assessed by our human evaluators. We therefore consider this driving segment to be instructive and use it as a fine-tuning dataset.  We first align the video frames with the control information in the CAN by time. Then, we use the speed, longitudinal \& lateral acceleration, and the sequence of steering angles of the vehicle in the first 0.1 seconds (recorded every 0.01 seconds) in the CAN as the control data of the vehicle, and the road surface geometry and the distance between the vehicle and the left and right roadside edges detected by the comma as the position data of the vehicle. We use these data as the description of the vehicle LKA, the pictures as the perception of the surrounding road surface, and the steering angle of the next control data from the CAN as the prediction of the correct decision, and send them to GPT 4-o through the API.

\begin{figure}[htbp]
  \centering
  \includegraphics[width=0.8\textwidth]{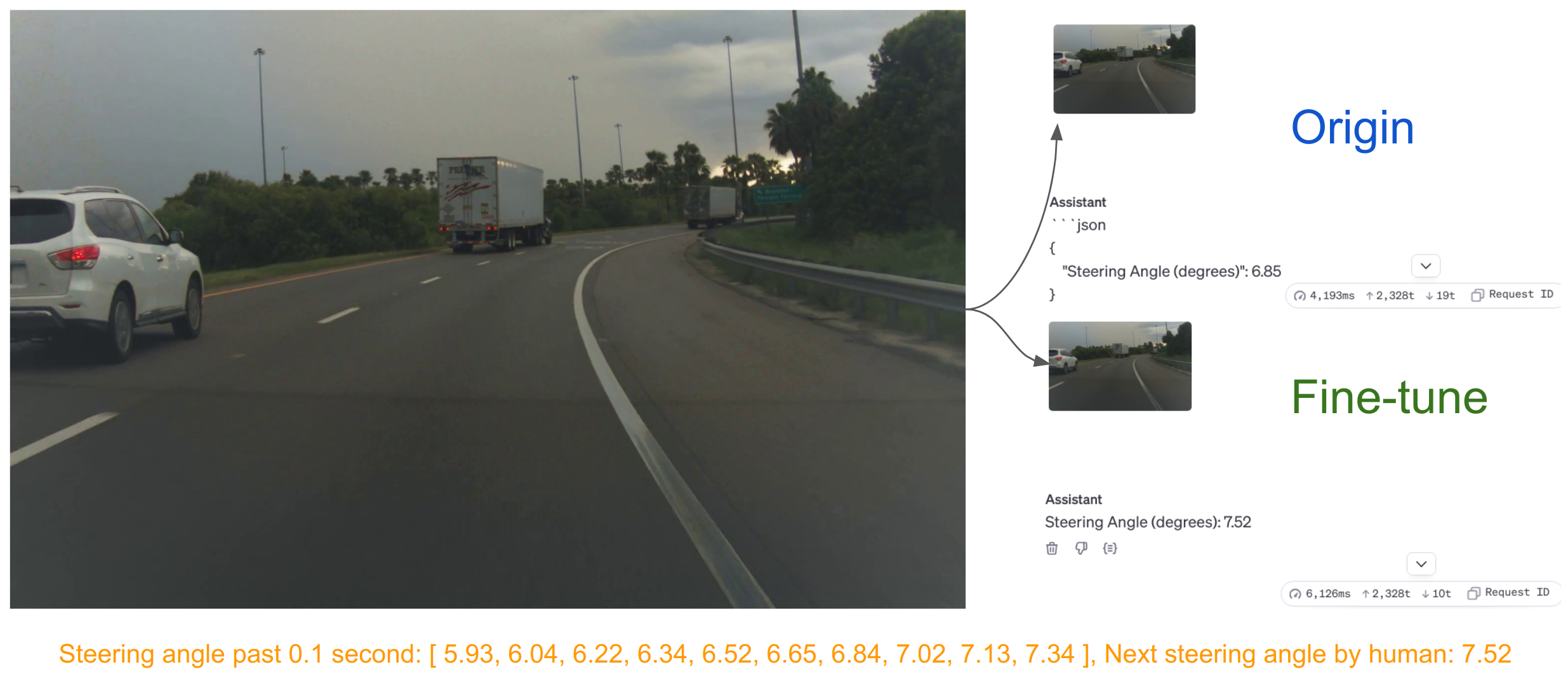}
  \caption{Performance Comparison between Original and Fine-tuned GPT Models}
  \label{fig:Comp1}
\end{figure}

\begin{figure}[htbp]
  \centering
  \includegraphics[width=0.8\textwidth]{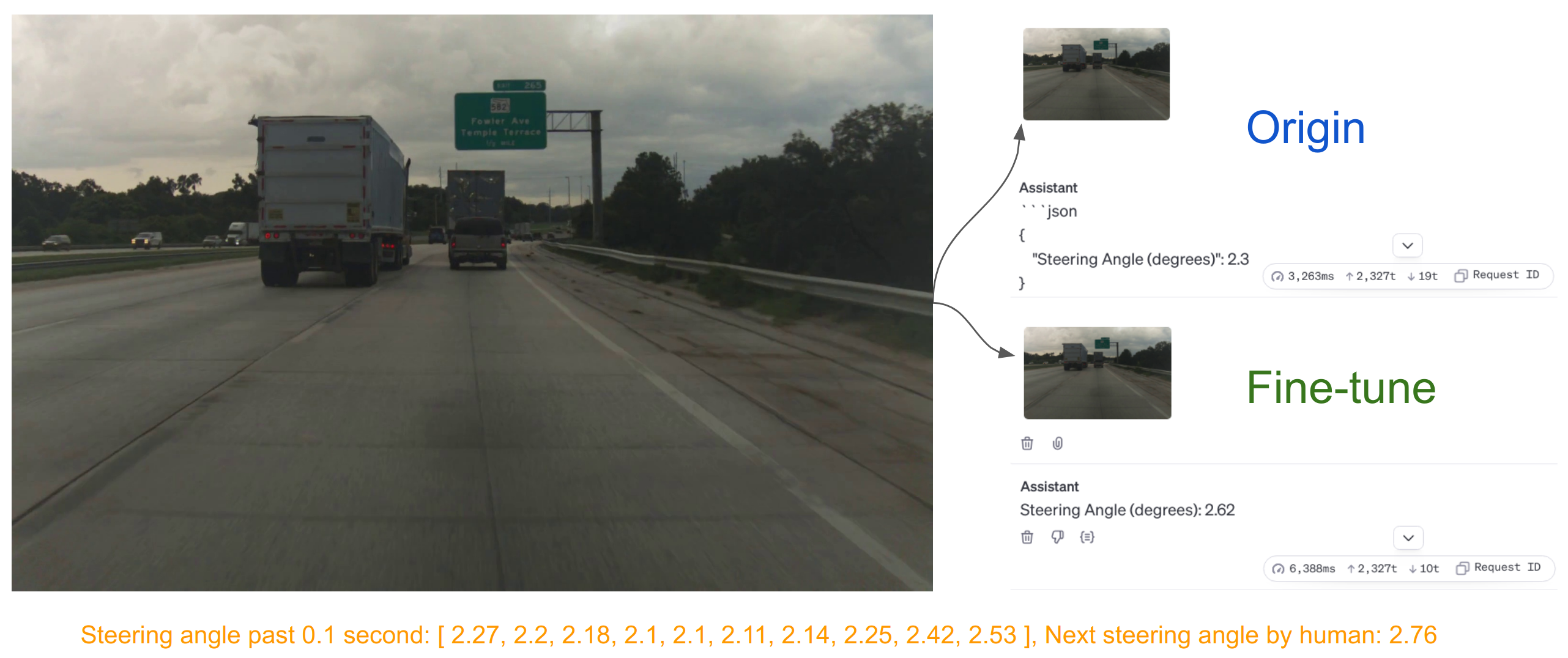}
  \caption{Advanced Scenario Comparison of Original and Fine-tuned GPT Models}
  \label{fig:Comp2}
\end{figure}

Through fine-tuning of more than 600+ combined picture-control information datasets, the results, shown in Fig~\ref{fig:loss}, training loss and validation loss, indicate that the model's output is very close to what a human driver would test next. In addition, we have designed specialized CoT prompts for iLKA to assist the model in making more rational decisions. The CoT process we designed is as follows \ref{appendix:lateral-cot}, we guide the visual language model to reason in a chain of thought from: recognizing itself and its surroundings - formulating the correct LKA lateral strategy - checking the vehicle's own control data to ensure safety - outputting the next planned control, this method to get the result. The fine-tuned GPT 4-o is able to make correct decisions and output steering angles that are very close to those of humans, which is a significant improvement over the origin GPT 4-o.  We compare the native GPT 4-o with the fine-tuned GPT model, as shown in Fig~\ref{fig:OriginVSFineTuned}, the output of the native GPT model is much lower than that of the actual human driver's control output. Fig~\ref{fig:Comp1} shows that during the cornering process, the native GPT cannot provide enough steering angle, which causes the vehicle to shift to the left significantly. If there are other vehicles in the left lane, as in the figure, this can be a serious safety hazard. Fig~\ref{fig:Comp2} shows another example that highlights the capability of the fine-tuned model, where the driver in the right-most lane is faced with large vehicles in the surrounding lanes, and the correct course of action is to keep the vehicle slightly to the right side of the road in order to avoid the potential risk. The fine-tuned model is able to give an adjustment that is closer to that of a human driver, whereas the adjustment given by the origin GPT model is still risky due to the under-steer angle.

In conclusion, we give a preliminary example of a planner with human flexibility in the iLKA framework, which not only provides more flexible decision planning, but also provides new ideas for personalized LKA and even ADAS systems. Integrating VLM and using appropriate CoT for guidance will greatly improve the safety and personalization of autonomous driving.

\textbf{Improvement on vision detection}

OpenLKA provides real-world driving videos with baseline lane detection from Openpilot, allowing researchers to benchmark and refine perception models under challenging conditions like faded markings, glare, and unclear road boundaries. It also includes trajectory predictions from vision-based models, enabling developers to compare their outputs, identify weaknesses, and improve system stability and reliability.

\textbf{Prediction of LKA Disengagement}

LKA systems, while enhancing vehicle safety in many scenarios, can occasionally lead to unintended drifts or even fail under specific conditions, creating potential safety risks. To address these challenges, our OpenLKA platform offers a comprehensive dataset designed to support advanced predictive modeling. By leveraging this extensive data, we aim to identify patterns and scenarios that may lead to LKA failures. This approach empowers developers and researchers to create more reliable algorithms, ultimately improving the safety and robustness of LKA systems.

\textbf{Implications for Cooperative Perception}

\begin{figure}[htbp]
  \centering
  \includegraphics[width=0.6\textwidth]{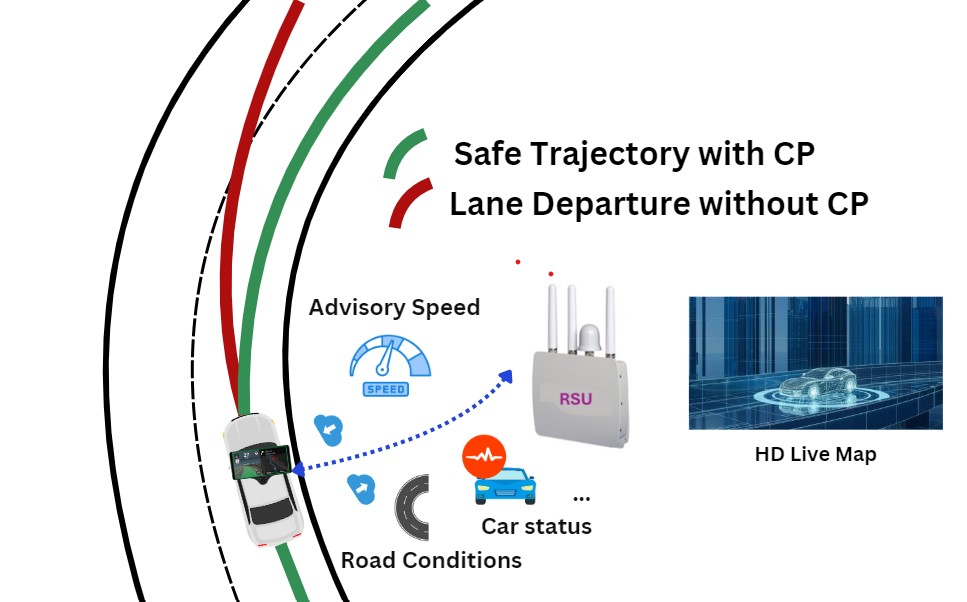}
  \caption{Architecture of Enhanced Cooperative Perception System Utilizing OpenLKA Dataset}
  \label{fig:CP_LKA}
\end{figure}

The integration of Cooperative Perception (CP) \cite{li2021cooperative} represents a pivotal advancement in addressing the limitations of current LKA systems, as underscored by the findings from the OpenLKA dataset. While traditional LKA systems rely heavily on onboard sensors to interpret their environment, their effectiveness is constrained by poor lane markings, adverse weather conditions, and challenging road geometries. CP offers a transformative solution by enabling vehicles and infrastructure to share perception data in real-time, thereby significantly enhancing the scope and accuracy of LKA functionalities.

The OpenLKA dataset serves as a critical resource for advancing CP systems, particularly in enhancing the capabilities of Roadside Units (RSUs) \cite{zhang2024evaluating, fang2024fedrsu}. By analyzing failure scenarios of existing LKA systems—such as degraded lane markings, adverse weather, and poor visibility—the dataset supports the development of RSU models that integrate sensor fusion (e.g., LiDAR, radar, high-definition maps) and real-time vehicle data. RSUs, in turn, can provide adaptive driving guidance tailored to diverse environments, issuing timely warnings and strategies during hazardous conditions like snowstorms or narrow oncoming traffic situations, ultimately improving safety and reducing collision risks \cite{zhang2024optimized, si2024bi}.

The potential applications inspired by the OpenLKA dataset extend beyond immediate improvements in LKA performance. By enabling RSUs and other CP infrastructure elements \cite{meng2024unified} to learn from diverse environmental challenges, the dataset can catalyze innovation that brings advanced cooperative perception capabilities to a wide range of vehicle platforms, including those with only basic ADAS features. These developments hold the promise of expanding the reach of autonomous safety technologies, spurring further research into robust, cooperative driving strategies, and ultimately paving the way for safer, more efficient roadways.


\newpage
\appendix

\section{Prompts for GPT4-o in annotation}
\label{appendix:prompts}

\begin{longtable}{|p{3cm}|p{13cm}|}
    \caption{\textbf{Prompt Structure}} \label{tab:prompt-structure} \\
    
    \hline
    \textbf{Component} & \textbf{Content} \\
    \hline
    
    System Prompts & 
    \textit{You are an AI language model specialized in transportation analysis}. Your task is to annotate given video frames using the provided labels. The device capturing these frames is installed in the center of the vehicle's windshield and is used to record the vehicle's driving data. \\
    \hline
    
    Task Description & 
    \begin{minipage}[t]{13cm}
    \textbf{Here're the labels and descriptions you need to follow:}\\[0.5em]
    \textbf{1. Lane Markings:}\\
    \hspace{1em}"0": "disappear: unable to see the lane markings" \\
    \hspace{1em}"1": "unclear markings: includes slight to moderate wear and tear, as well as obscurity due to weather conditions like rain or snow." \\
    \hspace{1em}"2": "intact markings: no wear and tear, all lane lines are clearly visible, and there is no risk of detection anomalies"\\[0.5em]
    
    \textbf{2. Weather:}\\
    \hspace{1em}"0": "rain and moisture"\\
    \hspace{1em}"1": "dusty: low visibility and yellow dust in air."\\
    \hspace{1em}"2": "clear weather"\\
    \hspace{1em}"3": "snowy"\\[0.5em]
    
    \textbf{3. Lighting:}\\
    \hspace{1em}"0": "dark"\\
    \hspace{1em}"1": "artificial lighting: street lights, etc."\\
    \hspace{1em}"2": "normal lighting: daylight, etc."\\[0.5em]

    \textbf{4. Traffic Congestion} (In the same direction of traffic):\\
    \hspace{1em}"0": "traffic jam: ego-vehicle stop in same place, without any movement"\\
    \hspace{1em}"1": "stop and go: higher number of stops on the way and high density of vehicles"\\
    \hspace{1em}"2": "congested flow: no stopping, but low speeds and little distance between each vehicle"\\
    \hspace{1em}"3": "free flow: appropriate speed and distance between cars"\\[0.5em]

    \textbf{5. Road Conditions:}\\
    \hspace{1em}"0": "large potholes, uneven or broken pavement"\\
    \hspace{1em}"1": "visible cracks, occasional small potholes"\\
    \hspace{1em}"2": "good condition, smooth and no visible cracks"\\[0.5em]

    \textbf{6. Driving Behavior:}\\
    \hspace{1em}"0": "Reckless driving: e.g., too close to the car in front, dangerous overtaking."\\
    \hspace{1em}"1": "Significant anomalies: e.g., abnormal deceleration, lane departure."\\
    \hspace{1em}"2": "Minor anomalies: e.g., slight lane deviation, minor deceleration."\\
    \hspace{1em}"3": "Normal driving, no anomalies"\\[0.5em]
    
    \textbf{7. Pedestrian Activity:}\\
    \hspace{1em}"0": "no pedestrians: 0 pedestrian."\\
    \hspace{1em}"1": "few pedestrians: less than 2 pedestrians, do not affect vehicle behavior."\\
    \hspace{1em}"2": "moderate activity: few pedestrians, vehicle sometimes decelerates."\\
    \hspace{1em}"3": "crowded: high interaction, making vehicle unable to drive properly."\\[0.5em]

    \end{minipage} \\
    \hline
    
\end{longtable}

\begin{longtable}{|p{3cm}|p{13cm}|}
    \hline
    \textbf{Component} & \textbf{Content} \\
    \hline

    Task Description & 
    \begin{minipage}[t]{13cm}  
    
    \textbf{8. Visibility:}\\
    \hspace{1em}"0": "poor visibility: severe impairment due to fog, heavy rain, snow, smoke, or obstructions."\\
    \hspace{1em}"1": "moderate visibility: partial impairment; objects are visible but not clearly."\\
    \hspace{1em}"2": "clear visibility: all objects are clearly visible."\\[0.5em]
    
    \textbf{9. Road Type:}\\
    \hspace{1em}"0": "two-lane road: one lane per direction."\\
    \hspace{1em}"1": "three-lane road: includes a two-way lane or variable center lane."\\
    \hspace{1em}"2": "four-lane road: two lanes per direction."\\
    \hspace{1em}"3": "multi-lane highway: three or more lanes in the same direction, with high speed limits and limited entrances and exits."\\
    \hspace{1em}"4": "roundabout: vehicles traveling around the traffic circle in a clockwise or counterclockwise direction."\\
    \hspace{1em}"5": "intersection: intersections or crossings where vehicles are required to follow traffic signals or markings."\\[0.5em]
    
    \textbf{10. Surrounding Vehicle:}\\
    \hspace{1em}"0": "no vehicles: no other vehicles around the ego-vehicle, including no visible vehicles in opposing traffic."\\
    \hspace{1em}"1": "low vehicle density: few vehicles, large distances between them, or opposing traffic with minimal vehicle presence."\\
    \hspace{1em}"2": "moderate vehicle density: a moderate number of vehicles in the same or opposing directions, with some distance between them."\\
    \hspace{1em}"3": "high vehicle density: vehicles are dense, small gaps, and heavy traffic, including opposing traffic that heavily interacts with the ego-vehicle."\\[0.5em]
    
    \textbf{11. Special Traffic Scenarios:}\\
    \hspace{1em}"0": "no special scenario"\\
    \hspace{1em}"1": "on-coming vehicle: ego-vehicle is currently in the leftmost lane and the on-coming car is less than 30m or passing by the ego-vehicle."\\
    \hspace{1em}"2": "emergency vehicle: ambulance, fire truck, police car."\\
    \hspace{1em}"3": "road construction or maintenance: construction zone, etc."\\
    \hspace{1em}"4": "obstacle on the road: debris, stalled vehicle, etc."\\
    \hspace{1em}"5": "animal crossing the road"\\
    \hspace{1em}"6": "accident scene"\\
    \end{minipage} \\
    \hline
    
    Instructions &
    \begin{minipage}[t]{13cm}
        \setlength{\parindent}{0pt}
        \textbf{1.} Analyze each provided video frame thoroughly using your transportation domain knowledge.\par\vspace{1em}
        
        \textbf{2.} Utilize the previous frame's annotation and image, as well as the next frame's image, to maintain labeling consistency, improve accuracy, and understand contextual changes.\par\vspace{1em}
        
        \textbf{3.} Let's think step by step to determine the correct labels for each category. Include your reasoning in the ``<description>'' field for each category.\par\vspace{1em}
        
        \textbf{4.} For the frame, output the final labels ``<label>'' in valid JSON format, ensuring proper use of quotation marks and commas.
    \end{minipage} \\[0.5em]
    \hline

    Format Description & 
    \begin{minipage}[t]{13cm}
    \textbf{Complete JSON Structure:}\\
    \ttfamily
    "frame": \{\\
    \hspace{1em}"Lane Markings": \{"description":"<description>", "label":"<label>"\},\\
    \hspace{1em}"Weather": \{"description":"<description>", "label":"<label>"\},\\
    \hspace{1em}"Lighting": \{"description":"<description>", "label":"<label>"\},\\
    \hspace{1em}"Traffic Congestion": \{"description":"<description>", "label":"<label>"\},\\
    \hspace{1em}"Road Conditions": \{"description":"<description>", "label":"<label>"\},\\
    \hspace{1em}"Driving Behavior": \{"description":"<description>", "label":"<label>"\},\\
    \hspace{1em}"Pedestrian Activity": \{"description":"<description>", "label":"<label>"\},\\
    \hspace{1em}"Visibility": \{"description":"<description>", "label":"<label>"\},\\
    \hspace{1em}"Road Type": \{"description":"<description>", "label":"<label>"\},\\
    \hspace{1em}"Surrounding Vehicle": \{"description":"<description>", "label":"<label>"\},\\
    \hspace{1em}"Special Traffic Scenarios": \{"description":"<description>", "label":"<label>"\}\\
    \} \\
    \end{minipage} \\
    \hline

    One-shot Example & 
    \begin{minipage}[t]{13cm}
    \textbf{Example Frame Analysis:}\\
    For the frame [\url{https://i.postimg.cc/g2LJjdg7/000000.jpg}], the output should be:\\[0.5em]
    \ttfamily
    "frame": \{\\
    \hspace{1em}"Lane Markings": \{"description":"The right lane marking is worn, some markings have disappeared, so the label is 1", "label":"1"\},\\
    \hspace{1em}"Weather": \{"description":"The light is good and everything is normal, so the label is 2.", "label":"2"\},\\
    \hspace{1em}"Lighting": \{"description":"It is sunny day, the lighting is normal, so the label is 2.", "label":"2"\},\\
    \hspace{1em}"Traffic Congestion": \{"description":"There's no car affecting the ego-vehicle velocity, it is free flow, so the label is 3.","label":"3"\},\\
    \hspace{1em}"Road Conditions": \{"description":"The road is even, but the right part has some small cracks, so the label is 1.", "label":"1"\},\\
    \hspace{1em}"Driving Behavior": \{"description":"It is normal driving, so the label is 3.","label":"3"\},\\
    \hspace{1em}"Pedestrian Activity": \{"description":"There is no pedestrian, so the label is 0.","label":"0"\},\\
    \hspace{1em}"Visibility": \{"description":"The visibility is clear, so the label is 2.", "label":"2"\},\\
    \hspace{1em}"Road Type": \{"description":"Is is a two-lane road, one is in ego direction and the other is on-coming direction, so the label is 0.","label":"0"\},\\
    \hspace{1em}"Surrounding Vehicle": \{"description":"There is an on-coming vehicle nearby, on the left of the ego-vehicle, but there's enough distance, so the label is 1.", "label":"1"\},\\
    \hspace{1em}"Special Traffic Scenarios": \{"description":"There is an on-coming vehicle on the left road passing by, it is an on-coming scenario, so the label is 1.","label":"1"\}\\
    \}\\[0.5em]
    \rmfamily
    I'll send you the current frame, the previous frame and the next frame to help you analyze the current frame better. Please give me the JSON result.
    \end{minipage} \\
    \hline
    
\end{longtable}


\section{Examples of prompts for GPT4-o in lateral decision making}
\label{appendix:lateral-cot}

\begin{longtable}{|p{2.5cm}|p{11cm}|}
    \caption{\textbf{CoT Prompts for Lateral Decision Making}} \label{tab:cot-lateral-prompts} \\
    \hline
    \textbf{Component} & \textbf{Content} \\
    \hline
    \textbf{System Prompts} & 
    \begin{minipage}[t]{10.5cm}
    \small
    \textit{You are an advanced AI driving assistant tasked with determining the optimal steering angle for an autonomous vehicle to maintain lane keeping and safety, analyzing the provided image and vehicle data. Make the right LKA decision and output the next steering angle.}\\[0.5em]
    \textbf{Image Reference:}\\
    \url{https://drive.google.com/file/d/1ccHUJm3GSegCGtEwbeZ6oPavv78ytRyf/view?usp=drivesdk}
    \normalsize
    \end{minipage}\\
    \hline
    \textbf{Vehicle Control Data} & 
    \begin{minipage}[t]{10.5cm}
    \small\texttt{%
    \{ \\ `Acceleration': 0.016, // Longitudinal acceleration (m/s\textasciicircum{}2)\\
    `Velocity': 30.782, // Vehicle speed (m/s)\\
    `Body pitch angle': -0.007, // Pitch angle (radians)\\
    `Body roll angle': 0.01, // Roll angle (radians)\\
    `Body yaw angle': -0.428, // Yaw angle (radians)\\
    `Body Roll': 0.013, // Body roll rate (radians/s)\\
    `Curvature': 0.0, // Current road curvature (1/m)\\
    `Distance to left laneline': -1.718, // Distance to left lane line (meters)\\
    `Distance to right laneline': 1.611, // Distance to right lane line (meters)\\
    `Lateral Acceleration': -0.128, // Lateral acceleration of the car\\
    `Steering angle past 0.1 second()': \\ 
    \hspace*{0.1em}[2.27, 2.2, 2.18, 2.1, 2.1, 2.11, 2.14,2.25, 2.42, 2.53] // History (degrees)\\
    \}  \\ }%
    \end{minipage}\\
    \hline
    \textbf{Step-by-Step Analysis (CoT)} & 
    \begin{minipage}[t]{10.5cm}
    \textbf{Let's think step by step:}\\[0.3em]
    \textbf{Step 1: Determining the External Environment}\\
    1.1 Road Type \& Geometry: Based on the image and curvature data, determine if the road is straight or curved.\\
    1.2 Current Road Position: Identify the ego vehicle's lane and position within that lane.\\
    1.3 Surrounding Vehicles: Analyze the image for surrounding vehicles (type, relative position).\\
    1.4 Pedestrians/Bicyclists: Identify any vulnerable road users and maintain safe distance.\\
    1.5 Weather: Infer conditions (rainy, foggy, etc.) from the image if possible, and adjust accordingly.\\[0.3em]
    \textbf{Step 2: Making the Right Strategy (to Ensure Safety)}\\
    2.1 Lane Keeping Strategy: Lean left, lean right, or adaptively center the lane based on obstacles or curves.\\[0.3em]
    \textbf{Step 3: Double-Checking Safety and Ego-Status}\\
    3.1 Road Geometry: Reconfirm the road curvature.\\
    3.2 Vehicle Dynamics: Consider speed, acceleration, body angles, and lateral acceleration.\\
    3.3 Current Position: Check lane line distances and marking validity.\\
    3.4 Steering History: Avoid abrupt or oscillatory maneuvers.\\[0.3em]
    \textbf{Step 4: Outputting the Steering Angle}\\
    4.1 Follow the chosen strategy.\\
    4.2 Steering Angle Calculation: Determine the optimal value.\\
    4.3 Output: Provide the final steering angle in degrees (positive = right turn, negative = left turn).
    \end{minipage}\\
    \hline
    \textbf{Final Output} & 
    \begin{minipage}[t]{10.5cm}
    \small
    \textit{Provide the steering angle in degrees, using the convention that positive values indicate a right turn and negative values indicate a left turn.}\\
    \textbf{Steering Angle (degrees):} [\textit{Your Answer Here}]
    \normalsize
    \end{minipage}\\
    \hline
\end{longtable}

\section{Road evaluation checklist for LKA system}
\label{appendix:checklist}

\begin{figure}[htbp]
  \centering
  \includegraphics[width=0.95\textwidth]{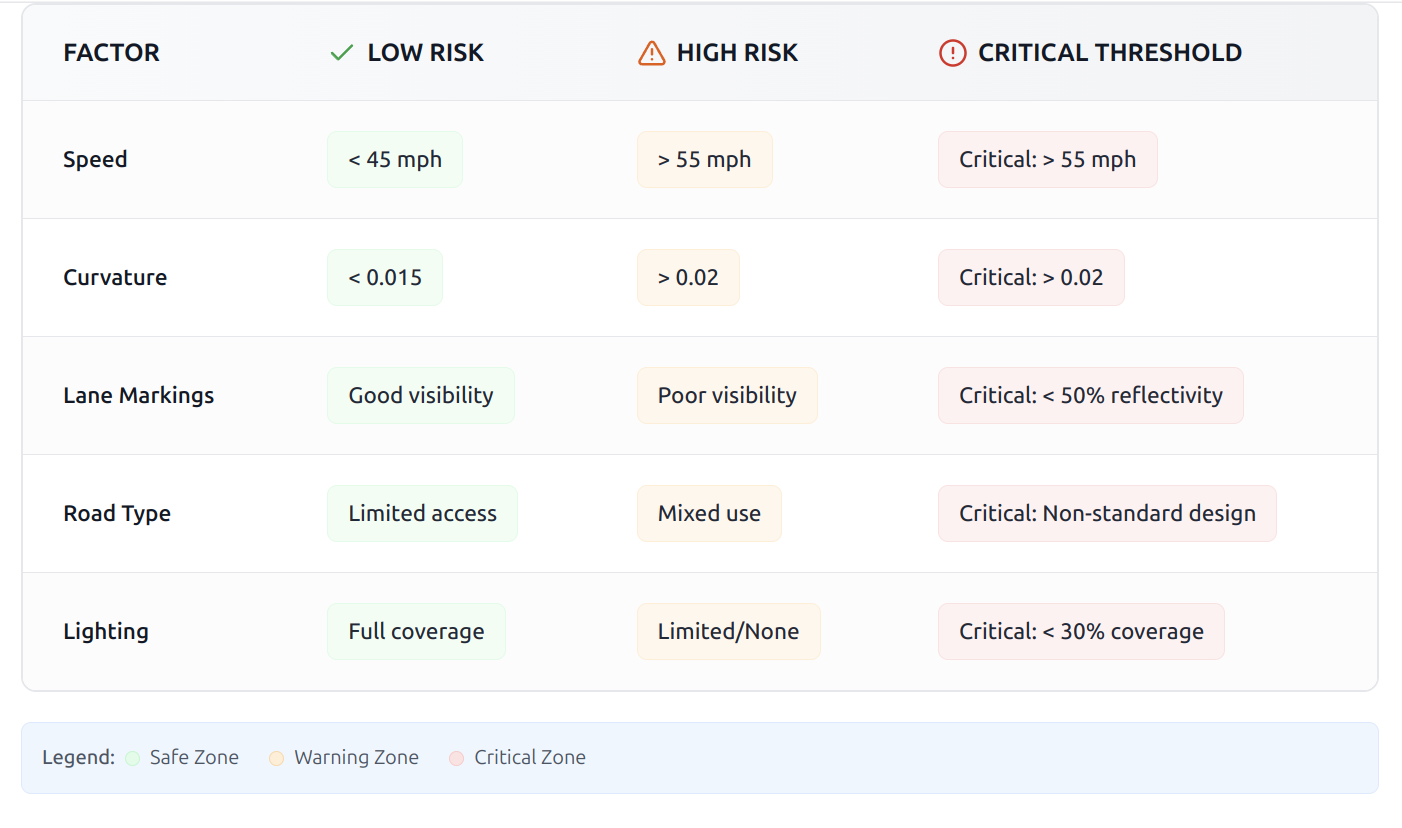}
  \caption{Checklist for Road Evaluation}
  \label{fig:dot_checklist}
\end{figure}

\bibliographystyle{unsrtnat} 
\bibliography{references}

\end{document}